%% file: main.tex
\documentclass[10pt,twocolumn,letterpaper]{article}

\RequirePackage[pagenumbers]{cvpr}   
\usepackage{pdfpages}  


\definecolor{cvprblue}{rgb}{0.21,0.49,0.74}
\usepackage[pagebackref,breaklinks,colorlinks,allcolors=cvprblue]{hyperref}

\title{Real-World Benchmarking and Synthetic Fine-Tuning of Monocular Metric Depth Estimation in Underwater Environments}

\author{Zijie Cai\\
University of Maryland, College Park\\
{\tt\small zai28@umd.edu}
\and
Christopher A. Metzler\\
University of Maryland, College Park\\
{\tt\small metzler.umd.edu}
}

\begin{document}


\maketitle

\input{sec/0_abstract}

\input{sec/1_intro}
\input{sec/2_related_work}

\input{sec/3_methods}
\input{sec/4_experiments}
\input{sec/5_discussion}
\input{sec/6_conclusion}
\input{sec/7_acknowledgments}

{
    \small
    \bibliographystyle{ieeenat_fullname}
    \bibliography{references}

}


\end{document}

%% file: sec/0_abstract.tex
\begin{abstract}
Monocular depth estimation has recently progressed beyond ordinal depth to provide metric depth predictions. However, its reliability in underwater environments remains limited due to light attenuation and scattering, color distortion, turbidity, and the lack of high-quality metric ground truth data. In this paper, we present a comprehensive benchmark of zero-shot and fine-tuned monocular metric depth estimation models on real-world underwater datasets with metric depth annotations, including FLSea and SQUID. We evaluated a diverse set of state-of-the-art Vision Foundation Models across a range of underwater conditions and depth ranges. Our results show that large-scale models trained on terrestrial data (real or synthetic) are effective in in-air settings, but perform poorly underwater due to significant domain shifts. To address this, we fine-tune Depth Anything V2 with a ViT-S backbone encoder on a synthetic underwater variant of the Hypersim dataset, which we simulated using a physically based underwater image formation model. Our fine-tuned model consistently improves performance across all benchmarks and outperforms baselines trained only on the clean in-air Hypersim dataset. This study presents a detailed evaluation and visualization of monocular metric depth estimation in underwater scenes, emphasizing the importance of domain adaptation and scale-aware supervision for achieving robust and generalizable metric depth predictions using foundation models in challenging environments.
\end{abstract}

%% file: sec/1_intro.tex
\section{Introduction}
\label{sec:intro}

Monocular depth estimation in complex underwater environments is critical for autonomous underwater vehicles (AUVs) for applications such as navigation \cite{lu2018survey, xanthidis2020navigation}, 3D mapping \cite{palomeras2018autonomous, wang2017underwater}, localization \cite{wang2017underwater, zhang2021fiducial}, object detection \cite{xu2023systematic, fayaz2022underwater}, and more. Unlike terrestrial robotics, underwater robotics systems lack dense depth sensing solutions \cite{sun2021review}. While LiDARs \cite{mcleod2013autonomous, dong2017lidar} and RGB-D cameras \cite{endres20133} are widely used above water, their deployment for underwater settings is severely limited by both hardware constraints and high cost \cite{zhou2021overview, zhuang2022underwater}. Acoustic Sonar systems are a more common alternative for marine applications \cite{rahman2019svin2, sun2021review, fallon2013relocating}, but their low spatial resolution due to constrained sensor elevation poses challenges for dense depth perception without additional sensor fusion \cite{qu2024z, kim2022underwater}. 

Recent advances in monocular depth estimation, particularly those based on vision foundation models using Vision Transformers (ViT) \cite{dosovitskiy2020image} and Dense Prediction Transformers (DPT) \cite{ranftl2021vision}, have achieved promising performance on in-air datasets like NYUv2 \cite{couprie2013indoor} and KITTI \cite{geiger2013vision}. These models typically leverage large-scale real and synthetic RGB-D data to learn powerful scene priors and for both relative and metric depth prediction from a single image \cite{yang2024depth, yang2024depthv2, nikolenko2021synthetic, godard2019digging}. However, their performance degrades significantly in underwater environments due to limited visibility caused by light scattering, turbidity, and wavelength-dependent attenuation \cite{li2019underwater, raveendran2021underwater, 8578801}. Furthermore, while traditional multi-view methods such as stereo matching \cite{chang2018pyramid} or SLAM \cite{taketomi2017visual, durrant2006simultaneous} can recover metric depth of the scene with geometric cues, they require multiple frames and consistent lighting, which are challenging to obtain underwater. In contrast, monocular methods require only a single frame for input, which offers much greater deployment flexibility \cite{ming2021deep, 10161471}. However, their zero-shot reliability in underwater settings remains a question \cite{zhang2025surveymonocularmetricdepth, bochkovskii2025depthprosharpmonocular}, not only due to the severe domain shifts in environments \cite{10048777}, but also due to the absence of geometric cues of the scene and the lack of accurate ground-truth metric depth data for supervision \cite{Zhang_2024_CVPR, 10.5555/1625275.1625630}.

In this work, we aim to address this domain gap by evaluating a diverse set of state-of-the-art monocular metric depth estimation models on two real-world underwater datasets with metric ground truth: FLSea \cite{randall2023flsea} and SQUID \cite{berman2018underwater}. Additionally, we explore the effectiveness of the underwater domain adaptation using a synthetically generated underwater dataset for fine-tuning to improve the model's generalization in underwater performance. 

\subsection*{Contributions}
Our paper presents a comprehensive benchmark and domain adaptation study of monocular metric depth estimation in underwater environments. Our main contributions are as follows: 

\begin{itemize}
\item We conduct an extensive zero-shot evaluation of six state-of-the-art monocular metric depth models with varying parameter sizes — including five general-purpose vision foundation models (Depth Anything V2 \cite{yang2024depthv2}, Metric3D V2 \cite{10638254}, UniDepth V2 \cite{piccinelli2025unidepthv2universalmonocularmetric}, ZoeDepth \cite{bhat2023zoedepthzeroshottransfercombining}, and Depth Pro \cite{bochkovskii2025depthprosharpmonocular}) and one underwater-specific method (UW-Depth) \cite{ebner2023metricallyscaledmonoculardepth} — on two real-world underwater datasets with metric ground truth: FLSea \cite{randall2023flsea} and SQUID \cite{berman2018underwater}.

\item We construct a large-scale synthetic underwater dataset by applying a physics-based underwater image formation model \cite{8578801} to the photorealistic Hypersim RGB-D dataset \cite{roberts:2021} of simulated indoor scenes with per-pixel ground truth labels, enabling low-cost, in-domain supervision for monocular metric depth estimation.  

\item We fine-tune Depth Anything V2 (ViT-S) \cite{yang2024depthv2} on our synthetically generated underwater version of the Hypersim dataset \cite{roberts:2021} and demonstrate consistent zero-shot monocular metric estimation improvements across all benchmarks compared to the baseline fine-tuned on the clean in-air Hypersim dataset \cite{roberts:2021}, highlighting the effectiveness of domain-adaptive synthetic training.

\item We provide both qualitative and quantitative comparisons across diverse underwater conditions using distinct subset scenes from the FLSea \cite{randall2023flsea} and SQUID \cite{berman2018underwater} datasets, analyzing model robustness and failure cases, and offering insights for future research in underwater perception.
\end{itemize}

%% file: sec/2_related_work.tex
\section{Related Work}
\label{sec:related_work}

\subsection{Monocular Metric Depth Estimation}
Monocular depth estimation has made significant progress in recent years with the increasing availability of large-scale RGB-D datasets \cite{geiger2013vision, couprie2013indoor} and the promising performance of transformer-based vision model architectures \cite{dosovitskiy2020image, ranftl2021vision}. These advancements have extended the development of general-purpose vision foundation models to predict metric depth of a scene in addition to relative depth from a single RGB image \cite{bhat2023zoedepthzeroshottransfercombining}. Recent state-of-the-art models such as Depth Anything V2 \cite{yang2024depthv2}, ZoeDepth \cite{bhat2023zoedepthzeroshottransfercombining}, and Metric3D V2 \cite{10638254} all adopt an encoder–decoder architecture \cite{DBLP:journals/corr/VaswaniSPUJGKP17}, where a ViT-based \cite{dosovitskiy2020image} or a convolutional-based \cite{Woo_2023_CVPR} encoder extracts high-level scene features and a task-specific decoder predicts dense depth maps \cite{ranftl2021vision}. Many of these models are trained using synthetic datasets with ground-truth per-pixel metric depth for supervision, such as Hypersim \cite{roberts:2021} and SceneNet \cite{McCormac_2017_ICCV}, alongside a large-scale, unlabeled real-world dataset for self-supervised learning with a teacher-student architecture through knowledge distillation \cite{hu2023teacherstudentarchitectureknowledgedistillation}. As a result from training, the models demonstrate robust performance for dense depth predictions across both indoor and outdoor in-air scenes by learning strong geometric and semantic priors \cite{yang2024depthv2}. 

\subsection{Depth Estimation Benchmarks}
Standard benchmarks for monocular depth estimation include NYU Depth v2 (indoor) \cite{couprie2013indoor} and KITTI (outdoor) \cite{geiger2013vision}, which provide dense metric ground truth for terrestrial scenes. However, few works provide an evaluation of state-of-the-art models in underwater conditions with real metric depth. Our work fills in this gap by evaluating six representative metric depth estimation models (Depth Anything V2 \cite{yang2024depthv2}, Metric3D V2 \cite{10638254}, UniDepth V2 \cite{piccinelli2025unidepthv2universalmonocularmetric}, ZoeDepth \cite{bhat2023zoedepthzeroshottransfercombining}, Depth Pro \cite{bochkovskii2025depthprosharpmonocular}, UW-Depth \cite{ebner2023metricallyscaledmonoculardepth}) on two real-world underwater datasets (FLSea \cite{randall2023flsea} and SQUID \cite{berman2018underwater}) with consistent qualitative visualizations and quantitative metrics (e.g., AbsRel, $\delta_1$) \cite{he2025distilldepthdistillationcreates, eigen2014depthmappredictionsingle}.

\subsection{Model Scaling and Inference Trade-offs}
Scaling vision models via larger ViT backbones (e.g., ViT-S, ViT-B, ViT-L, ViT-G) generally improves depth prediction accuracy \cite{chen2021outperform, zhuang2022gsam, steiner2021augreg, dosovitskiy2020image}. However, larger models also lead to higher latency in training and inference, and more extensive memory demands, which limit their deployment for real-time applications \cite{10834497}. Evaluating models across sizes helps identify trade-offs between inference efficiency and accuracy, which is critical for embedded systems such as autonomous underwater vehicles (AUVs), which are often hardware-constrained \cite{10161471}.

\subsection{Underwater Depth Estimation}
Underwater scenes pose unique challenges for depth estimation due to complex light interactions with water. The visibility in underwater imagery is often limited, caused by issues such as wavelength-dependent attenuation, backscatter, turbidity, and non-uniform illumination \cite{10.1145/3578516, Akkaynak_2019_CVPR}. While acoustic sensors such as sonar are commonly used for underwater range sensing, they suffer from low spatial resolution due to limited elevation coverage \cite{qu2024z}. Monocular metric depth estimation methods designed specifically for underwater environments, such as UW-Depth \cite{ebner2023metricallyscaledmonoculardepth}, are often trained on real-world underwater datasets. However, the scarcity of large-scale, high-quality metric ground-truth data limits model complexity and generalization. As a result, these methods often adopt lightweight encoder–decoder architectures optimized for speed rather than dense accuracy, and their performance tends to degrade outside their training domain \cite{mehta2022mobilevitlightweightgeneralpurposemobilefriendly}.

\subsection{Synthetic Data for Model Training}
To address the lack of ground truth underwater data, recent research has turned to synthetic data generation using physics-based rendering \cite{10406819}. The underwater image formation model \cite{8578801, 10.1145/3578516} simulates critical optical effects such as attenuation and backscatter, allowing in-air RGB-D datasets to be transformed into realistic underwater imagery. This approach has been adopted widely in image restoration and enhancement tasks, where obtaining ground-truth color-corrected underwater images is practically infeasible \cite{9570386}. By allowing supervised training with per-pixel metric depth and controlled variation in water conditions, this synthetic-to-real strategy serves as an effective tool for domain adaptation and model pretraining across various underwater vision tasks \cite{10048777}.

\subsection{Domain Adaptation in Vision}
Domain adaptation methods aim to improve model generalization across data distributions by aligning features, styles, or learned representations for a new domain \cite{7078994}. In underwater vision, where the domain shift from terrestrial imagery is severe, prior work has employed strategies such as CycleGAN-based image translation for underwater image enhancement tasks \cite{DBLP:journals/corr/ZhuPIE17, 9570386}. The authors of \textit{Atlantis: Enabling Underwater Depth Estimation with Stable Diffusion} \cite{Zhang_2024_CVPR} also propose a pipeline for generating underwater imagery for depth estimation with a diffusion model \cite{10081412} and a control-net-based approach for training underwater relative depth estimation models \cite{zhang2023addingconditionalcontroltexttoimage}. However, such approaches introduce significant overhead in data preparation and add complexity to multi-stage pipelines, which slows down both training and inference for the model. 

In contrast, we adopt a forward supervised adaptation strategy \cite{Motiian_2017_ICCV} by fine-tuning a general-purpose monocular depth model (Depth Anything V2 \cite{yang2024depthv2}) on a synthetic underwater dataset generated using a physics-based rendering pipeline to generalize its performance for the underwater metric depth domain. Our supervised fine-tuning strategy (SFT) is lightweight, requiring no auxiliary networks for synthetic data generation, and freezes the early encoder layers to retain the strong pre-trained scene understanding priors while adapting the decoder and late encoder layers to underwater-specific image statistics with labeled data \cite{kumar2022finetuningdistortpretrainedfeatures}.

%% file: sec/3_methods.tex
\section{Methods}
\label{sec:methods}

\subsection{Overview}
\label{sec:methods-overview}

This section outlines our methodology for benchmarking and fine-tuning monocular metric depth estimation in underwater environments. Our pipeline consists of:

\begin{enumerate}
    \item \textbf{Zero-shot evaluation of existing models:} We benchmark a diverse set of six state-of-the-art monocular metric depth estimation models, five of which are general-purpose foundation models and the other one is underwater-specific, on two real-world underwater datasets (FLSea \cite{randall2023flsea} and SQUID \cite{berman2018underwater}).
    
    \item \textbf{Synthetic dataset generation:} To address the scarcity of real-world underwater metric ground truth, we create a synthetic underwater dataset by applying a physics-based underwater image formation model to an existing clean in-air synthetic RGB-D dataset (Hypersim \cite{roberts:2021}). This simulates various underwater imaging conditions while preserving high-quality per-pixel metric depth \cite{8578801}.
    
    \item \textbf{Domain adaptation via supervised fine-tuning:} We fine-tune a pretrained ordinal depth foundation model (Depth Anything V2 \cite{yang2024depthv2}) using the synthetic underwater Hypersim \cite{roberts:2021} dataset to adapt the model for predicting underwater metric depth. This approach enables the model to learn underwater-specific visual cues while retaining the generalization ability from pretraining.
\end{enumerate}

\subsection{Benchmark Models}
\label{sec:benchmark-models}

We evaluate six state-of-the-art monocular metric depth estimation models, of which five are general-purpose vision foundation models and one is an underwater-specific approach. All models are assessed in a zero-shot setting using publicly available pretrained weights. For in-domain adaptation, we select Depth Anything V2 (ViT-S) \cite{yang2024depthv2} as our fine-tuning baseline and explore its performance on the real-world underwater benchmarks. 

\begin{itemize}
    \item \textbf{Depth Anything V2 \cite{yang2024depthv2}}: A transformer-based depth foundation model originally trained for general-purpose ordinal depth estimation, which can also be fine-tuned for in-domain metric depth estimation. It employs a ViT \cite{dosovitskiy2020image} encoder and DPT \cite{ranftl2021vision} decoder and supports multiple encoder scales (ViT-S, ViT-B, ViT-L). We evaluate all three variants that the authors fine-tuned on the clean Hypersim dataset and use ViT-S for our synthetic training.

    \item \textbf{Depth Pro \cite{bochkovskii2025depthprosharpmonocular}}: A foundation model for zero-shot metric monocular depth and focal length estimation optimized for efficient inference on resource-constrained platforms to provide high-resolution depth maps with unparalleled sharpness and high frequency details. 

    \item \textbf{Metric3D V2 \cite{10638254}}: A geometric foundation model for zero-shot metric depth and surface normal estimation from a single image with the proposed canonical camera space transformation module to address metric ambiguity. 

    \item \textbf{UniDepth V2 \cite{piccinelli2025unidepthv2universalmonocularmetric}}: A universal transformer model for zero-shot monocular metric depth estimation that predicts 3D point clouds directly from a single image without requiring camera intrinsics. It employs a self-promptable camera module and geometric invariance losses to predict a dense camera representation for conditioning depth features and enhancing generalization across domains.

    \item \textbf{ZoeDepth \cite{bhat2023zoedepthzeroshottransfercombining}}: A scale-aware model that combines ordinal and metric depth cues using a lightweight bin-adjusted head and latent classifier. It is trained on multiple relative and metric datasets to achieve strong zero-shot generalization and scale consistency across domains. Their flagship model, ZoeD-M12-NK, is used for our experiments. 

    \item \textbf{UW-Depth \cite{ebner2023metricallyscaledmonoculardepth}}: A lightweight, domain-specific model trained on underwater datasets such as FLSea \cite{randall2023flsea}. It incorporates sparse feature priors to mitigate scale ambiguity for underwater metric depth estimation and is optimized for real-time inference on embedded systems.
    
\end{itemize}

\subsection{Synthetic Underwater Dataset Generation}
To enable supervised fine-tuning in underwater conditions, we generate a synthetic underwater training set based on the Hypersim \cite{roberts:2021} RGB-D dataset, which provides a dense set of photorealistic indoor scenes and corresponding ground-truth metric depth. 

We apply the simplified underwater image formation model for ambient illumination, assuming the camera response is equivalent to the delta function \cite{8578801}. This formulation accounts for wavelength-dependent attenuation and backscattering. Specifically, we use the wideband approximation of their model, expressed as \cite{1315078, 6104148, 5567108}:

\begin{equation}
I_c = J_c \cdot e^{-\beta_c z} + B^{\infty}_c \cdot (1 - e^{-\beta_c z}),
\label{eq:akkaynak}
\end{equation}

where $I_c$ is the observed underwater intensity in channel $c \in \{R, G, B\}$, $J_c$ is the clear scene radiance, $\beta_c$ is the wideband attenuation coefficient for channel $c$, $B^{\infty}_c$ is the wideband veiling light (backscatter at infinity), and $z$ is the range (depth) from the camera. This model captures both the exponential attenuation of the direct signal and the accumulation of backscattered light.

We simulate multiple Jerlov water types from open (I, II, III) to coastal ocean classes (1C to 9C) with varying $\beta_c$ and $B^{\infty}_c$ to represent different optical properties and beam absorption levels of ocean water \cite{Solonenko:15}. The resulting dataset consists of paired underwater RGB images and their clean metric depth maps. All images preserve the original pixel alignment and camera intrinsics from Hypersim \cite{roberts:2021}.

\begin{figure}[ht]
    \centering
    \setlength{\tabcolsep}{1pt}
    \renewcommand{\arraystretch}{0.5}
    \begin{tabular}{ccccc}
        \scriptsize RGB & \scriptsize Depth & \scriptsize Type I & \scriptsize Type II & \scriptsize Type III \\
        \includegraphics[width=0.19\linewidth]{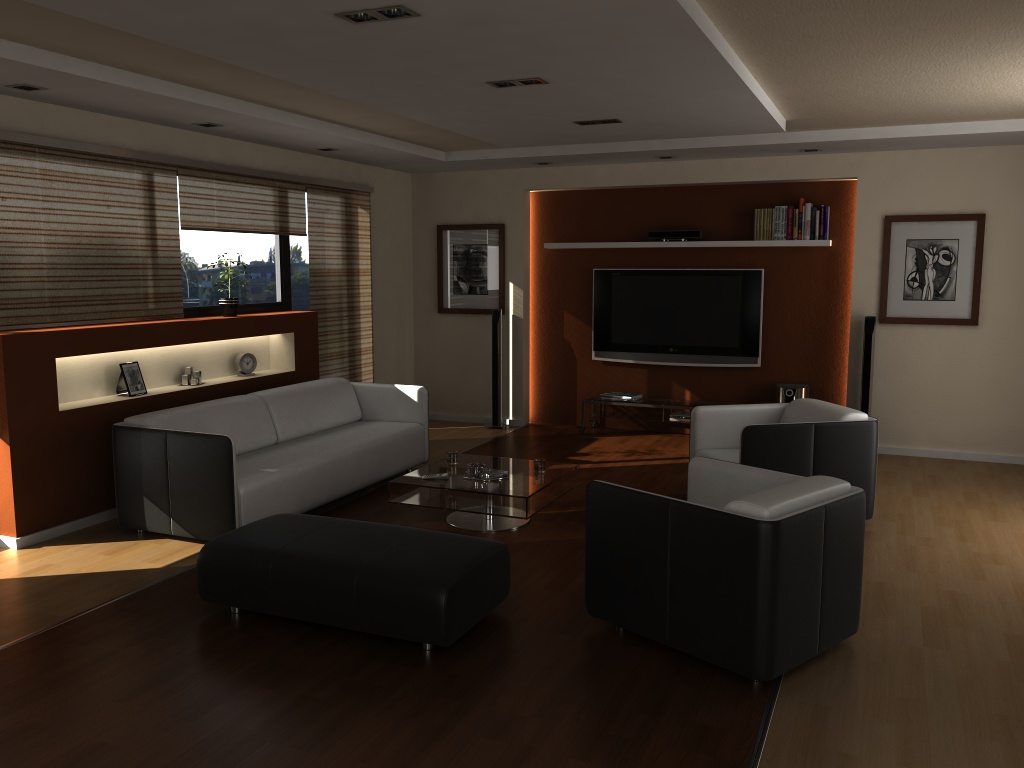} &
        \includegraphics[width=0.19\linewidth]{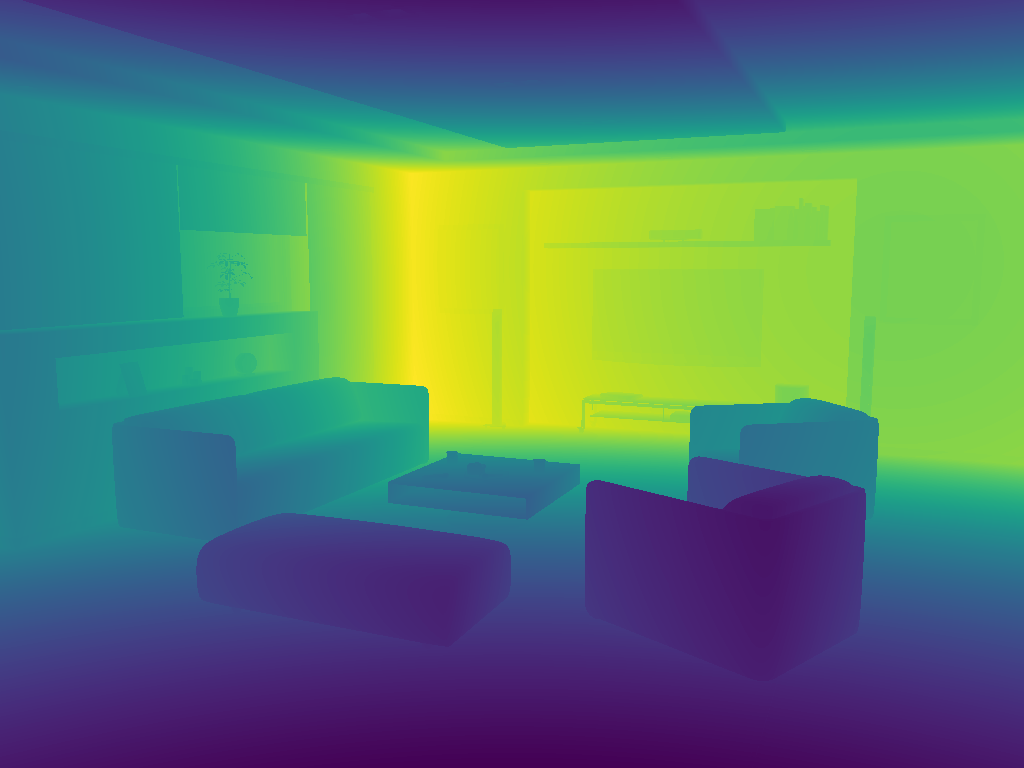} &
        \includegraphics[width=0.19\linewidth]{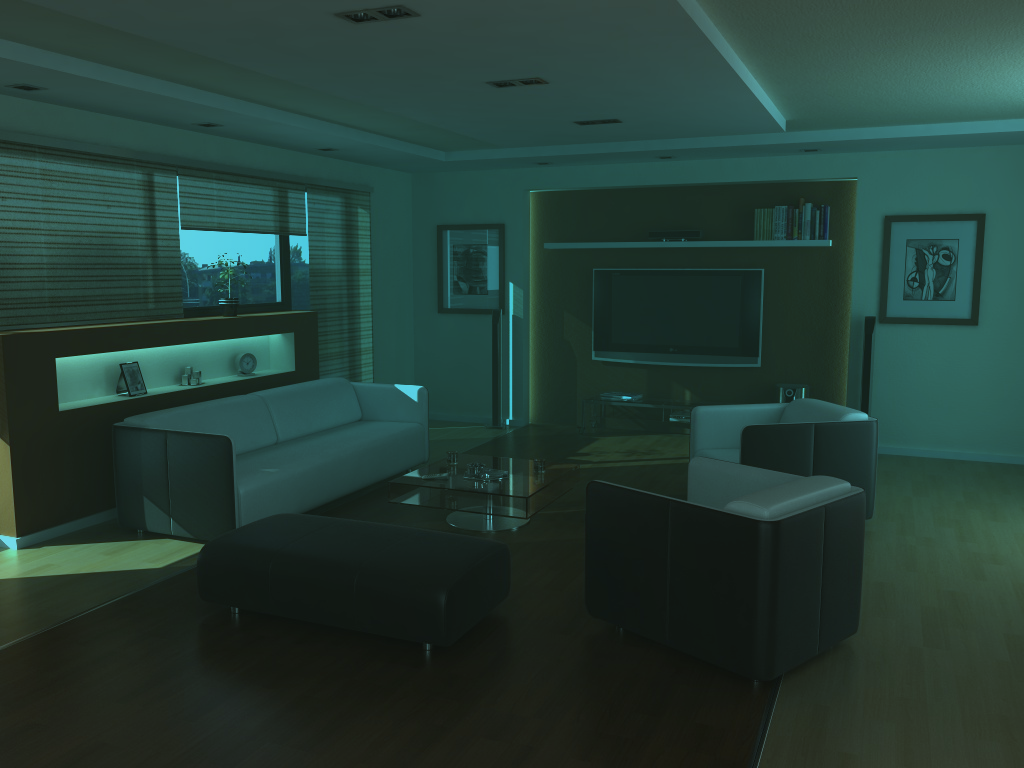} &
        \includegraphics[width=0.19\linewidth]{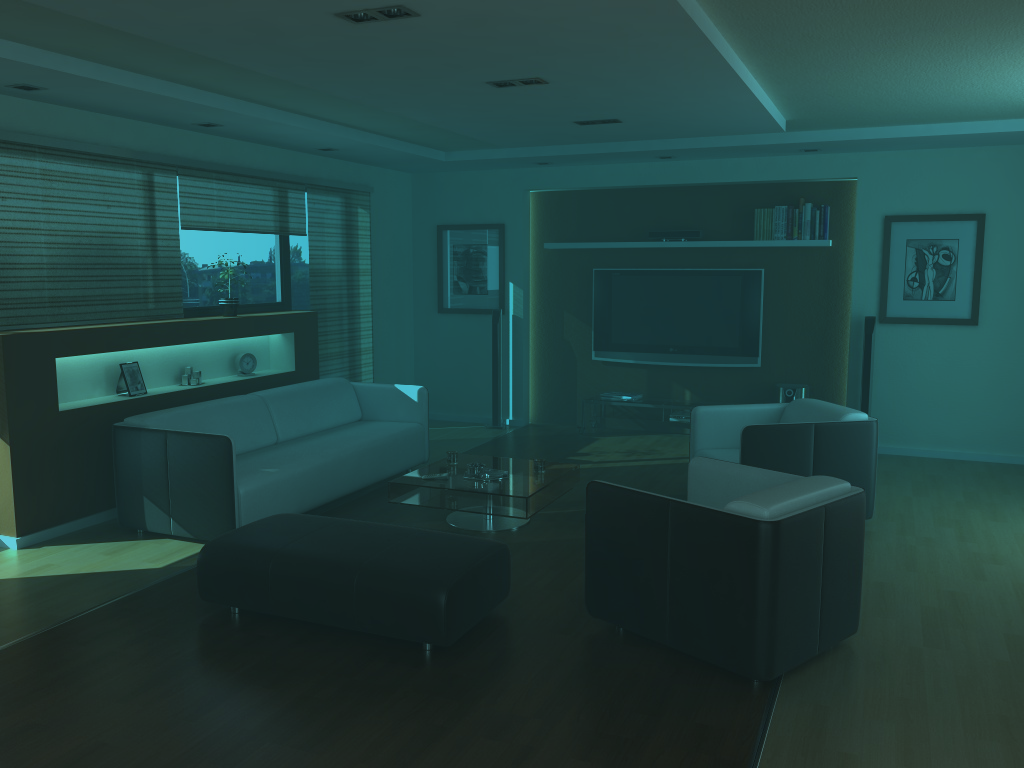} &
        \includegraphics[width=0.19\linewidth]{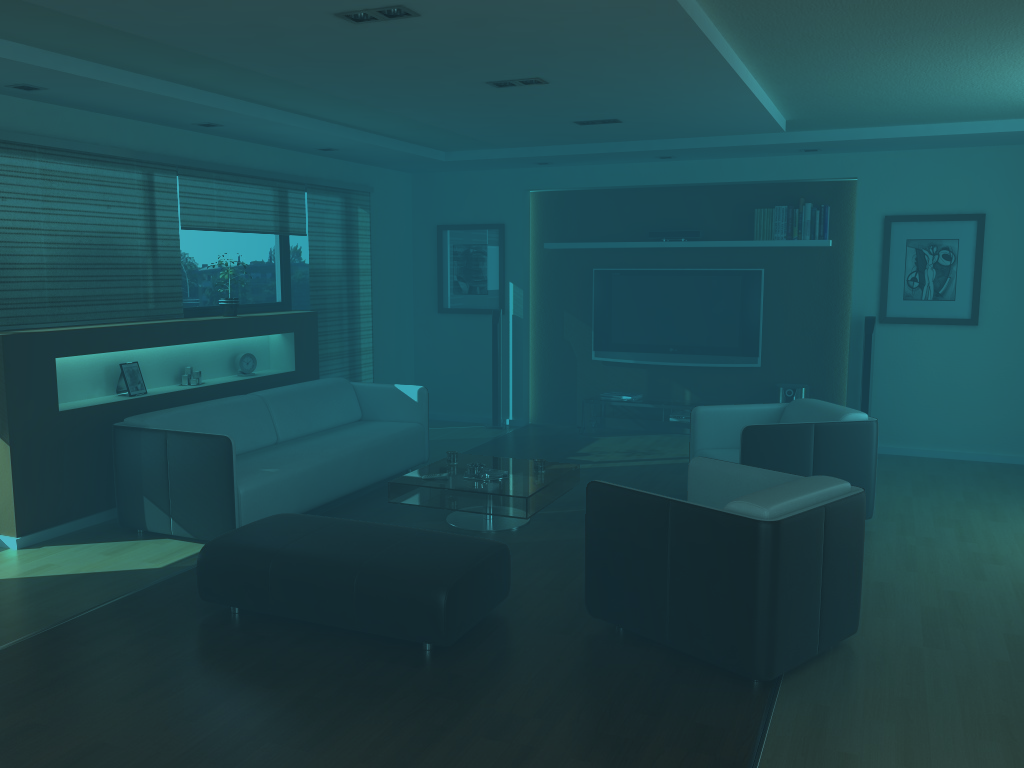} \\[4pt]

        \scriptsize Type 1C & \scriptsize Type 3C & \scriptsize Type 5C & \scriptsize Type 7C & \scriptsize Type 9C \\
        \includegraphics[width=0.19\linewidth]{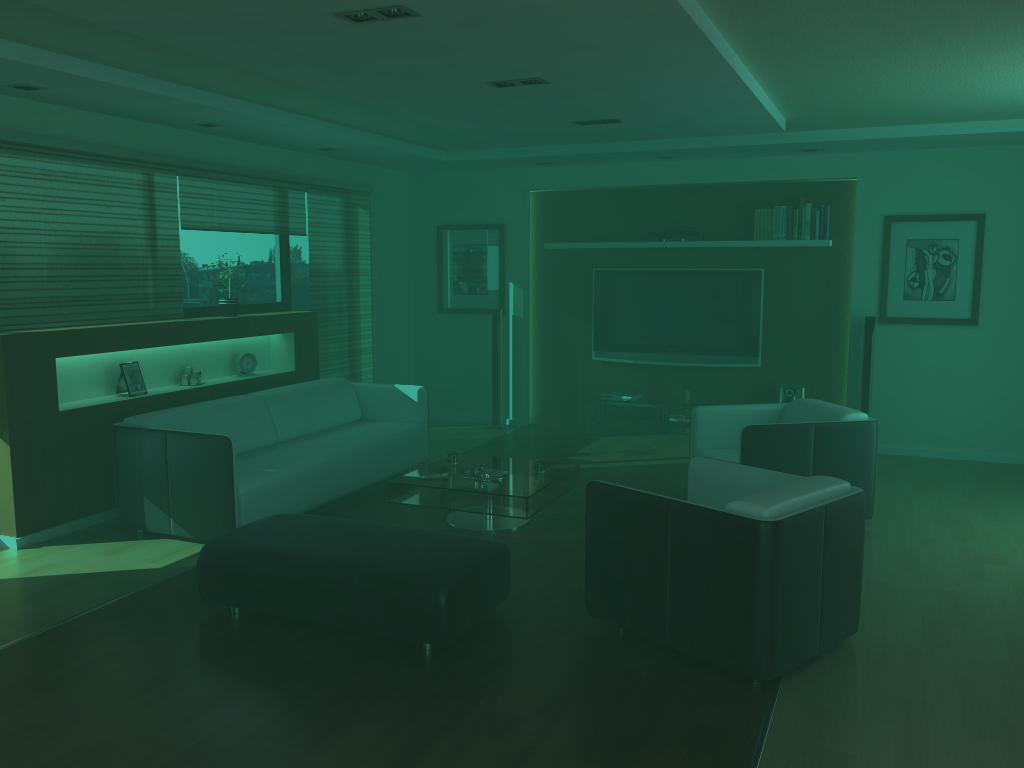} &
        \includegraphics[width=0.19\linewidth]{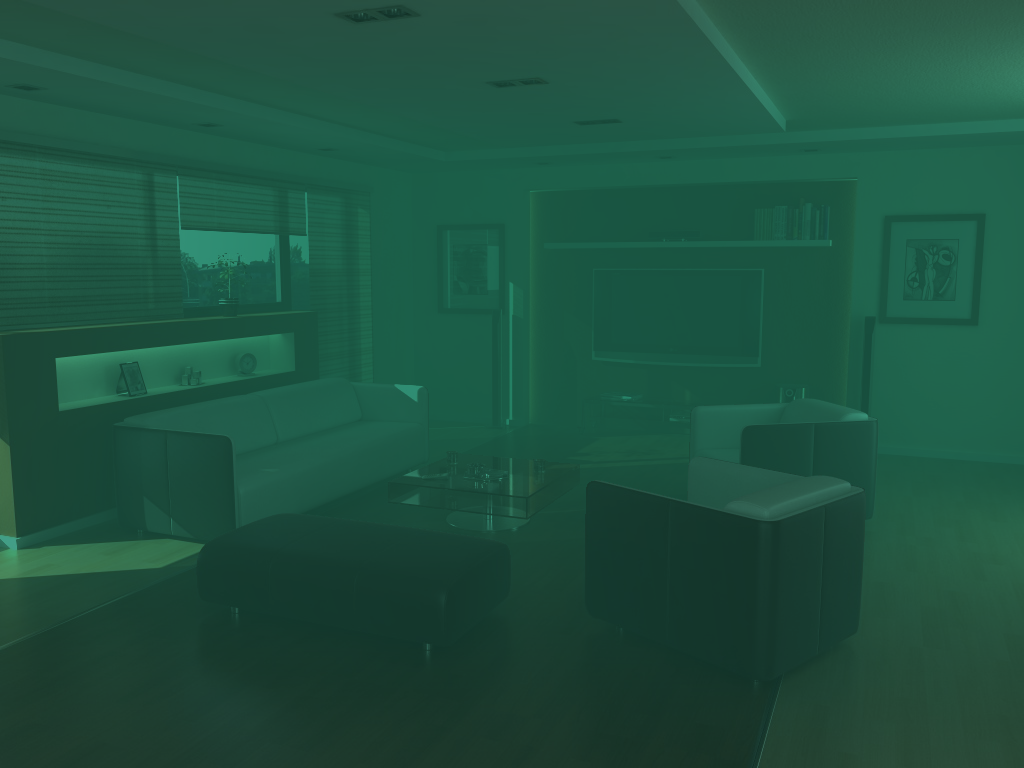} &
        \includegraphics[width=0.19\linewidth]{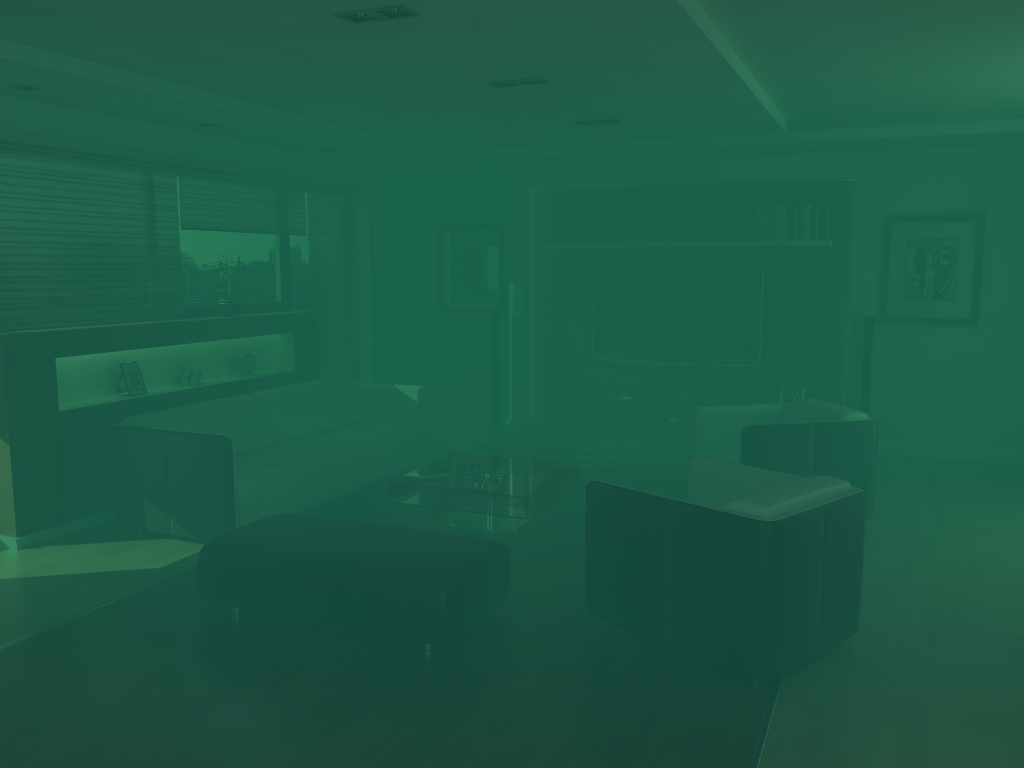} &
        \includegraphics[width=0.19\linewidth]{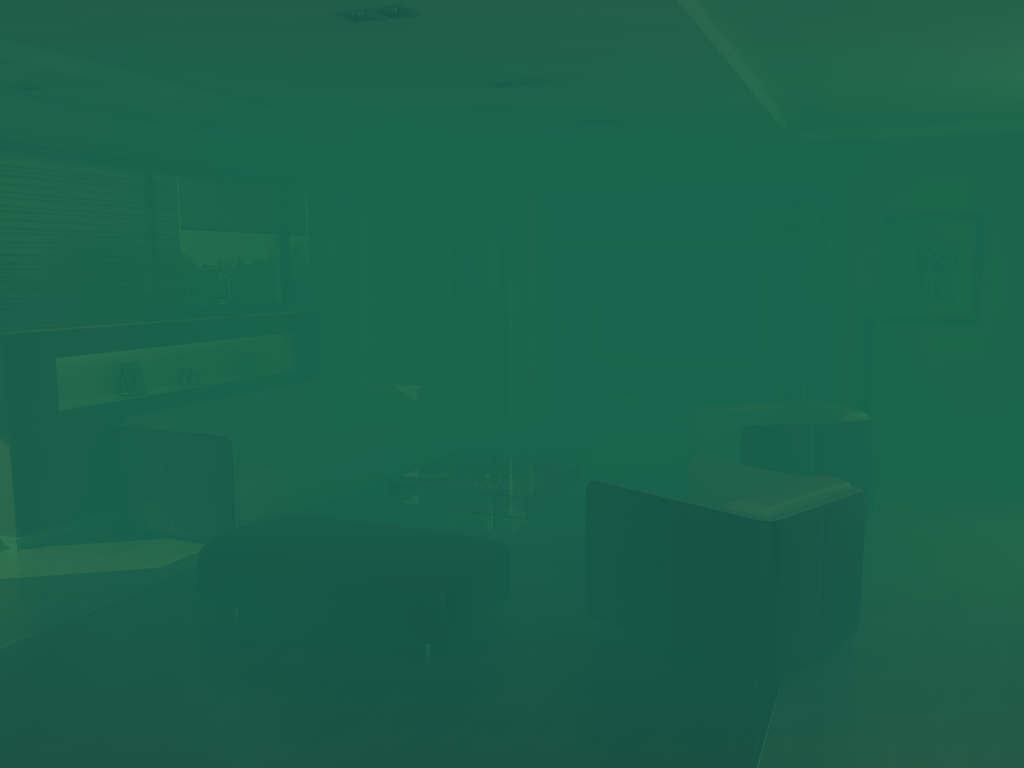} &
        \includegraphics[width=0.19\linewidth]{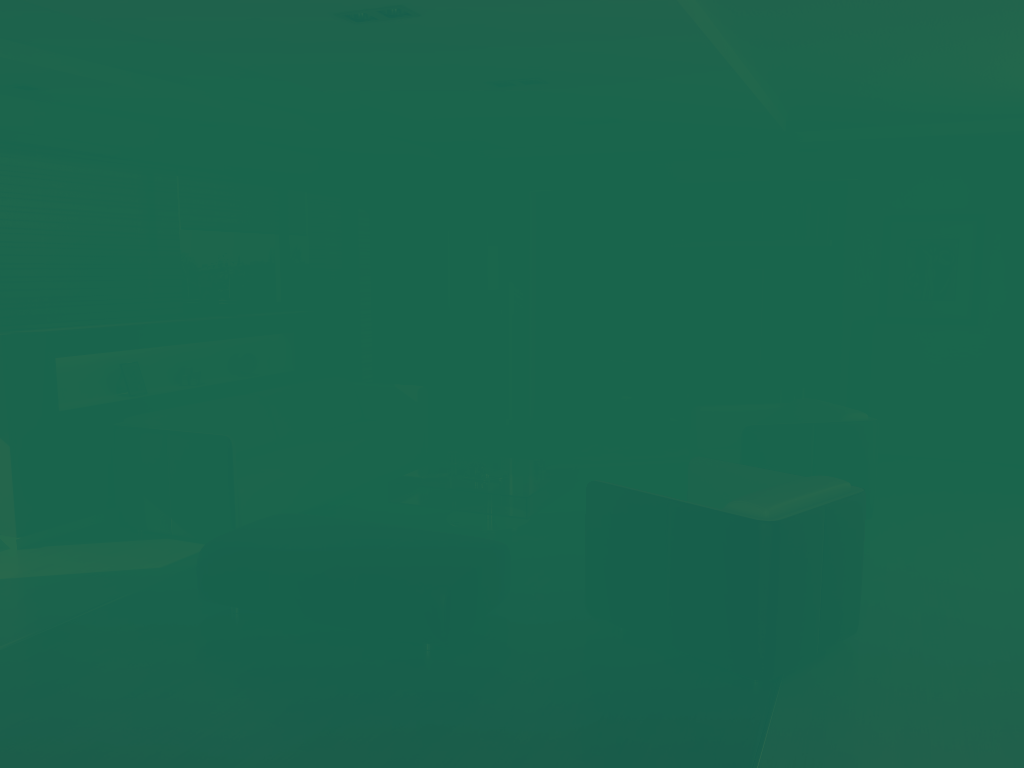} \\
    \end{tabular}
    \caption{
    Examples from our synthetic underwater dataset. Top row: clean RGB image, ground-truth depth, and Jerlov open-ocean classes (I, II, III). Bottom row: simulations with increasing attenuation in coastal ocean classes (1C to 9C), representing progressively turbid conditions \cite{Solonenko:15}.
    }
    \label{fig:synthetic_conditions}
\end{figure}

\subsection{Fine-Tuning Depth Anything on Synthetic Data}
We use a supervised fine-tuning strategy using our synthetic underwater dataset in order to adapt the Depth Anything V2 \cite{yang2024depthv2} model to the unique underwater imagery domain \cite{10048777}. This dataset includes varying visual distortions typical of underwater scenes, which each image is one of the Jerlov water classes we saw earlier. Our goal is to enable the model to generalize to the real underwater environment while retaining the strong priors learned from large-scale terrestrial RGB-D and unlabeled data. 

We use the ViT-S variant of Depth Anything V2 \cite{yang2024depthv2} as our baseline and initialize it with the official relative depth checkpoint. To retain previously learned features, we freeze the first half of the Vision Transformer encoder and allow only the remaining encoder layers and DPT-style decoder to update during training, which is conducted in a supervised regression setting with the following configuration \cite{ranftl2021vision, 10.1007/978-3-319-49409-8_34}:

\begin{itemize}
    \item \textbf{Optimizer:} AdamW \cite{loshchilov2019decoupledweightdecayregularization} with weight decay of $10^{-2}$
    \item \textbf{Learning rate:} $5 \times 10^{-6}$ with cosine annealing scheduler \cite{loshchilov2017sgdrstochasticgradientdescent}
    \item \textbf{Warm-up:} Linear warm-up for the first 4 epochs \cite{kalra2024warmuplearningrateunderlying}
    \item \textbf{Epochs:} 20 total
    \item \textbf{Batch size:} 4 (constrained by GPU memory)
    \item \textbf{Loss function:} SiLogLoss \cite{eigen2014depthmappredictionsingle} — a scale-invariant logarithmic loss, commonly used for metric depth regression.
    \item \textbf{Max depth:} 20 meters (used for scaling model output)
\end{itemize}

We train the model using image–depth pairs at a resolution of $518 \times 518$. To enhance robustness for underwater domain adaptation, we apply physically consistent color augmentations, including random illumination changes to simulate under- and overexposure in real-world underwater settings, as well as grayscale conversion to encourage the model to focus on structural cues rather than color variations caused by water conditions \cite{9745315, 10.1007/978-3-030-87237-3_25}.

Our supervised fine-tuning strategy enables the model to learn underwater-specific visual cues while retaining general-purpose scene priors from pretraining. By leveraging synthetic data aligned with underwater image formation physics, the fine-tuned model achieves more reliable metric depth estimation in challenging underwater environments.

\subsection{Evaluation Datasets}
We evaluate all models on two real-world underwater datasets containing RGB images paired with metric ground-truth depth:

\begin{itemize}
    \item \textbf{FLSea \cite{randall2023flsea}:} A large-scale dataset collected in controlled underwater environments using diver-operated cameras and photogrammetry software (Agisoft Metashape) to generate ground-truth metric depth via SFM \cite{ozyesil2017surveystructuremotion}. We focus on six key subsets from two scenes:

    \begin{itemize}
        \item \textit{Canyon:} \texttt{u\_canyon} and \texttt{flatiron}, totaling 5,369 images. These subsets capture natural rocky reef structures at water depths of 4–7 meters.
        \item \textit{Red Sea:} \texttt{big\_dice\_loop}, \texttt{cross\_pyramid\_loop}, \texttt{coral\_table\_loop}, and \texttt{sub\_pier}, totaling 6,919 images. These scenes include both natural structures and large man-made objects such as coral tables, piers, and concrete blocks, with water depths ranging from 3–8 meters.
    \end{itemize}
    
    \item \textbf{SQUID \cite{berman2018underwater}:} A smaller but more challenging dataset with longer depth ranges and larger areas of featureless background, consisting of 57 stereo image pairs with metric ground-truth depth computed from stereo triangulation \cite{chang2018pyramid, 1211491}. The dataset covers four subset scenes collected across different marine environments in Israel:
    
    \begin{itemize}
        \item \textit{Tropical Red Sea:} 
        \begin{itemize}
            \item \texttt{Katzaa} — coral reef, 10–15 meters deep.
            \item \texttt{Satil} — shipwreck site, 20–30 meters deep.
        \end{itemize}
        \item \textit{Temperate Mediterranean Sea:}
        \begin{itemize}
            \item \texttt{Nachsolim} — rocky reef, 3–6 meters deep.
            \item \texttt{Michmoret} — rocky reef, 10–12 meters deep.
        \end{itemize}
    \end{itemize}
\end{itemize}

All predictions are rescaled to match the dataset-specific depth units. Quantitative evaluations are performed using standard depth estimation metrics, as detailed in Section~\ref{sec:experiments}.

%% file: sec/4_experiments.tex
\section{Experiments}
\label{sec:experiments}

\subsection{Evaluation Setup}
We evaluate all models on two real-world underwater datasets: FLSea \cite{randall2023flsea} and SQUID \cite{berman2018underwater}. For each model, we use official pre-trained weights and perform zero-shot inference unless otherwise specified. Depth Anything V2 \cite{yang2024depthv2} is additionally fine-tuned on our synthetic underwater dataset. When applicable, we evaluate model variants using different ViT encoder sizes (ViT-S, ViT-B, ViT-L).

\subsection{Metrics}
We report standard monocular depth estimation metrics, widely used in prior work \cite{he2025distilldepthdistillationcreates, eigen2014depthmappredictionsingle}:
\begin{itemize}
    \item \textbf{AbsRel} (Absolute Relative Error): $\frac{1}{|T|}\sum_{i \in T} \frac{|d_i - \hat{d}_i|}{d_i}$
    \item \boldmath$\delta_1$\unboldmath: Percentage of predictions satisfying $\delta = \max(\frac{d_i}{\hat{d}_i}, \frac{\hat{d}_i}{d_i}) < 1.25$
\end{itemize}

Here, $d_i$ and $\hat{d}_i$ represent the ground-truth and predicted depths, respectively. Metrics are computed only on valid (non-zero) ground-truth pixels and follow the dataset-specific evaluation protocols.

\subsection{Zero-Shot Benchmarking Results}
We first evaluate all models with official pre-trained weights to compare zero-shot performance to underwater scenes. \Cref{tab:combined_results} summarizes the results across both datasets. 

Notably, several models come with specific limitations in the zero-shot setting. \textbf{Metric3D V2} \cite{10638254} performance varies with input resolution, and we exclude its ViT-L variant due to hardware constraints during inference. \textbf{Depth Anything V2} \cite{yang2024depthv2}, while trained primarily on ordinal depth, provides competitive results but requires further in-domain only fine-tuning to predict metric depth. \textbf{UW-Depth} \cite{ebner2023metricallyscaledmonoculardepth} is trained on 10 subsets of the FLSea \cite{randall2023flsea} dataset and two held-out test sets (u\_canyon and sub\_pier). To avoid evaluation with data leakage, we omit its quantitative results on FLSea \cite{randall2023flsea}.

\begin{table*}[ht]
\centering
\begin{tabular}{lcc|cc|cc}
\toprule
\textbf{Model} & 
\multicolumn{2}{c|}{\textbf{FLSea-Canyon}} & 
\multicolumn{2}{c|}{\textbf{FLSea-Red Sea}} & 
\multicolumn{2}{c}{\textbf{SQUID}} \\
& AbsRel $\downarrow$ & $\delta_1$ $\uparrow$
& AbsRel $\downarrow$ & $\delta_1$ $\uparrow$
& AbsRel $\downarrow$ & $\delta_1$ $\uparrow$ \\
\midrule
ZoeDepth & 1.5907 & 0.2345 & 1.3335 & 0.2109 & 1.3214 & 0.0968 \\
Metric3D V2 (ViT-S)$^{\dagger}$ & 1.5331 & 0.0967 & 0.8130 & 0.2136 & 1.3059 & 0.1680 \\
Depth Pro & 0.9858 & 0.1557 & 0.3888 & 0.3772 & 3.2185 & 0.1678 \\
UW-Depth$^{\dagger}$ & -- & -- & -- & -- & 0.4948 & 0.3446 \\
Depth Anything V2 (ViT-S)$^{\dagger}$ & 0.3576 & 0.4463 & 0.2569 & 0.4722 & 0.5242 & 0.2054 \\
Depth Anything V2 (ViT-B)$^{\dagger}$ & 0.2447 & 0.5696 & 0.2471 & 0.4301 & 0.4495 & 0.2649 \\
Depth Anything V2 (ViT-L)$^{\dagger}$ & 0.2269 & 0.6363 & 0.2307 & 0.4812 & 0.3390 & 0.2896 \\
UniDepth V2 (ViT-S) & 0.2233 & 0.6763 & 0.1524 & 0.8122 & 0.4012 & 0.4789 \\
UniDepth V2 (ViT-B) & 0.1276 & 0.8844 & 0.1045 & 0.9167 & 0.3638 & 0.4725 \\
UniDepth V2 (ViT-L) & \textbf{0.1156} & \textbf{0.9109} & \textbf{0.0932} & \textbf{0.9439} & \textbf{0.3222} & \textbf{0.5201} \\
\bottomrule
\end{tabular}
\caption{
Zero-shot performance comparison across FLSea-Canyon \cite{randall2023flsea}, FLSea-Red Sea \cite{randall2023flsea}, and SQUID \cite{berman2018underwater} datasets. 
$^{\dagger}$Models marked with this symbol are not strictly zero-shot for metric depth. Bold indicates best performance.
}
\label{tab:combined_results}
\end{table*}

\subsection{Effect of Synthetic Fine-Tuning}
We fine-tune Depth Anything V2 (ViT-S) \cite{yang2024depthv2} on our synthetic underwater dataset and compare its performance to the baseline fine-tuned for in-domain metric depth prediction on the clean Hypersim \cite{roberts:2021} dataset with the maximum depth scale of 20 meters. This experiment evaluates the effectiveness of forward domain adaptation using synthetic underwater data. Quantitative improvements are summarized in \Cref{tab:fine_tune_results}.


\begin{table*}[ht]
\centering
\begin{tabular}{lcc|cc|cc}
\toprule
\textbf{Model (ViT-S)} & 
\multicolumn{2}{c|}{\textbf{FLSea-Canyon}} & 
\multicolumn{2}{c|}{\textbf{FLSea-Red Sea}} & 
\multicolumn{2}{c}{\textbf{SQUID}} \\
& AbsRel $\downarrow$ & $\delta_1$ $\uparrow$
& AbsRel $\downarrow$ & $\delta_1$ $\uparrow$
& AbsRel $\downarrow$ & $\delta_1$ $\uparrow$ \\
\midrule
Baseline DA V2   & \textbf{0.3576} & 0.4463 & 0.2569 & 0.4722 & 0.5242 & 0.2054 \\

Fine-tuned DA V2 (Ours) & 0.3620 & \textbf{0.4683} & \textbf{0.2266} & \textbf{0.6170} & \textbf{0.4465} & \textbf{0.3204} \\
\bottomrule
\end{tabular}
\caption{Performance of Depth Anything V2 (ViT-S) \cite{yang2024depthv2} before and after fine-tuning on synthetic underwater data across FLSea-Canyon \cite{randall2023flsea}, FLSea-Red Sea \cite{randall2023flsea}, and SQUID \cite{berman2018underwater} subsets. Bold indicates best performance.}
\label{tab:fine_tune_results}
\end{table*}

\subsection{Qualitative Results}
\begin{figure*}[ht]
    \centering
    \renewcommand{\arraystretch}{1.0}
    \setlength{\tabcolsep}{1pt}
    \scriptsize

    \begin{tabular}{cccccccc}
        \multicolumn{8}{c}{\textbf{FLSea: Canyon}} \\
        \textbf{RGB} & \textbf{GT} & \textbf{UW-Depth$^{\dagger}$} & \textbf{Metric3D-L$^{\dagger}$} & 
        \textbf{DA V2-L$^{\dagger}$} & \textbf{Depth Pro} & \textbf{UniDepth-L} & \textbf{ZoeDepth} \\
        \includegraphics[width=0.115\textwidth]{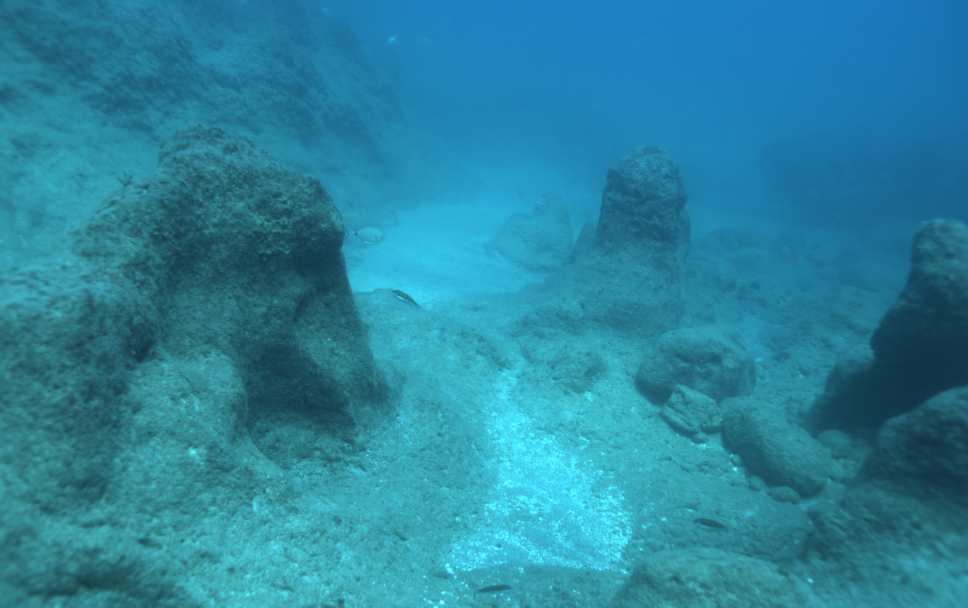} &
        \includegraphics[width=0.115\textwidth]{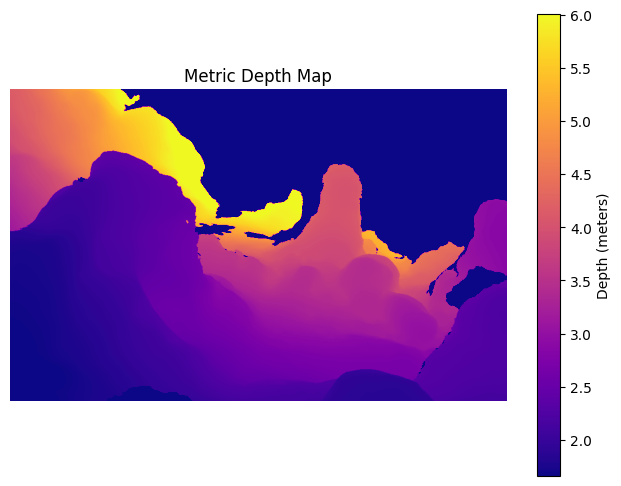} &
        \includegraphics[width=0.115\textwidth]{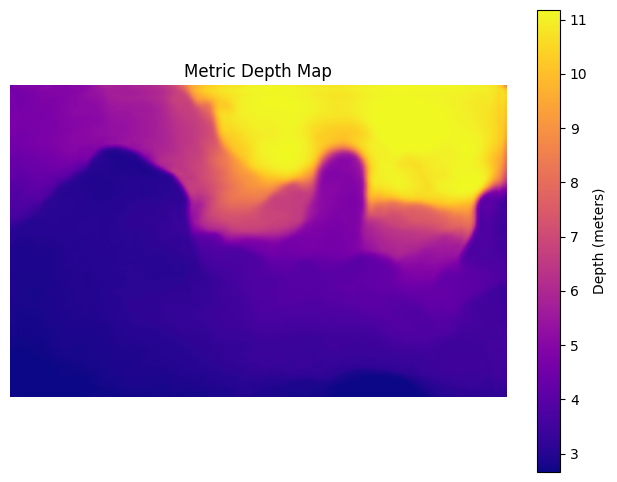} &
        \includegraphics[width=0.115\textwidth]{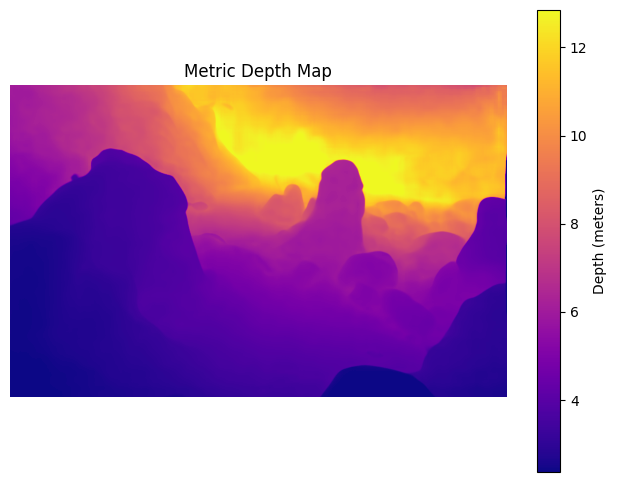} &
        \includegraphics[width=0.115\textwidth]{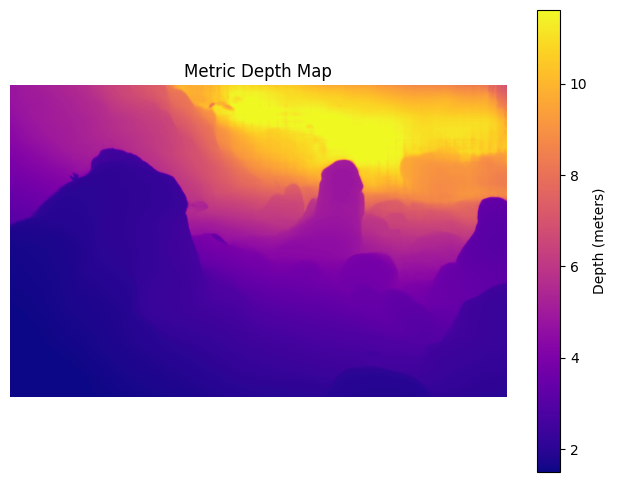} &
        \includegraphics[width=0.115\textwidth]{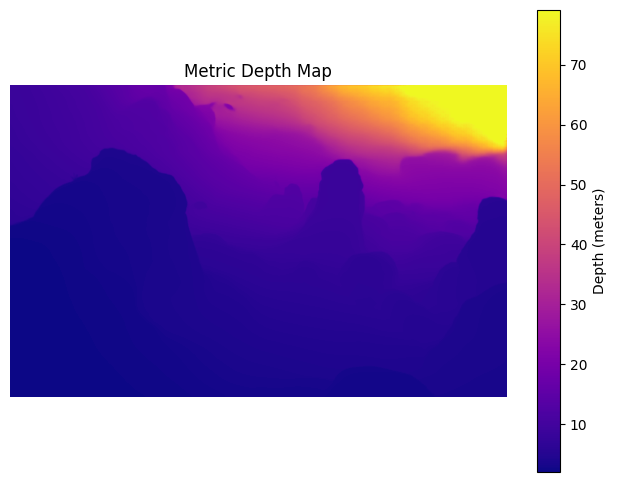} &
        \includegraphics[width=0.115\textwidth]{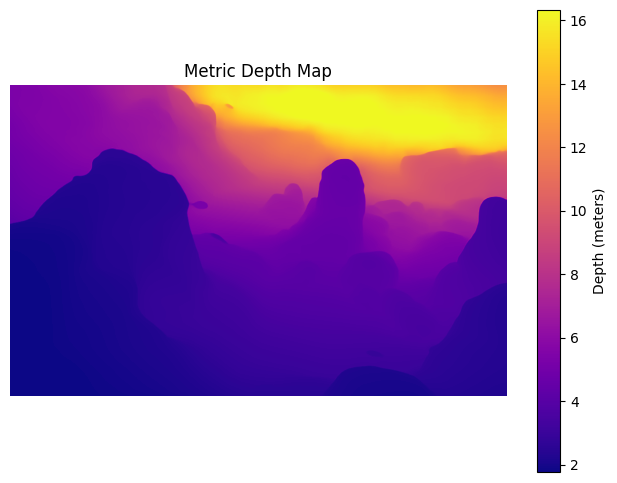} &
        \includegraphics[width=0.115\textwidth]{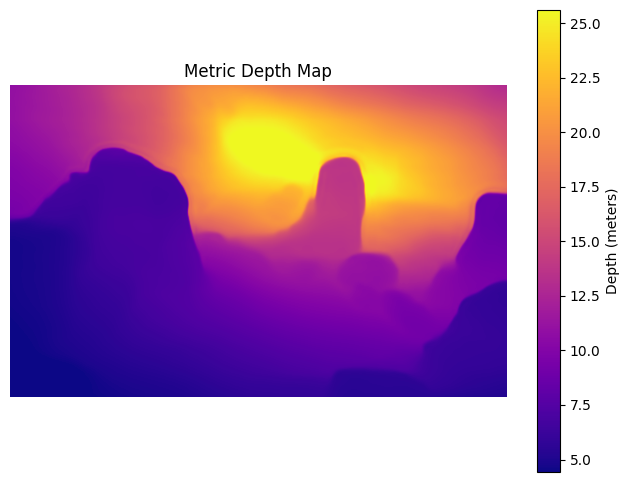} \\
    \end{tabular}

    \vspace{6pt}

    \begin{tabular}{cccccccc}
        \multicolumn{8}{c}{\textbf{FLSea: Red Sea}} \\
        \textbf{RGB} & \textbf{GT} & \textbf{UW-Depth$^{\dagger}$} & \textbf{Metric3D-L$^{\dagger}$} & 
        \textbf{DA V2-L$^{\dagger}$} & \textbf{Depth Pro} & \textbf{UniDepth-L} & \textbf{ZoeDepth} \\
        \includegraphics[width=0.115\textwidth]{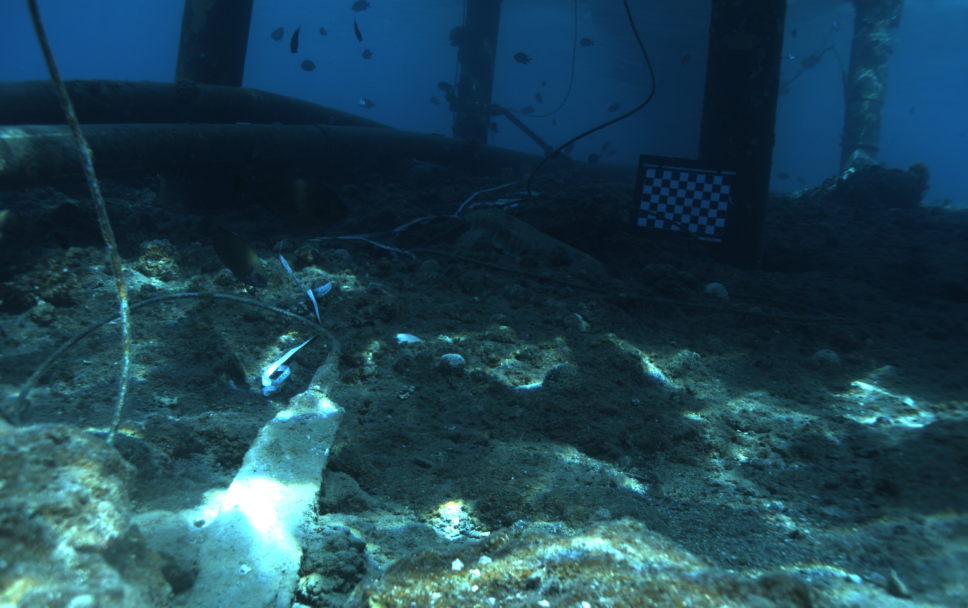} &
        \includegraphics[width=0.115\textwidth]{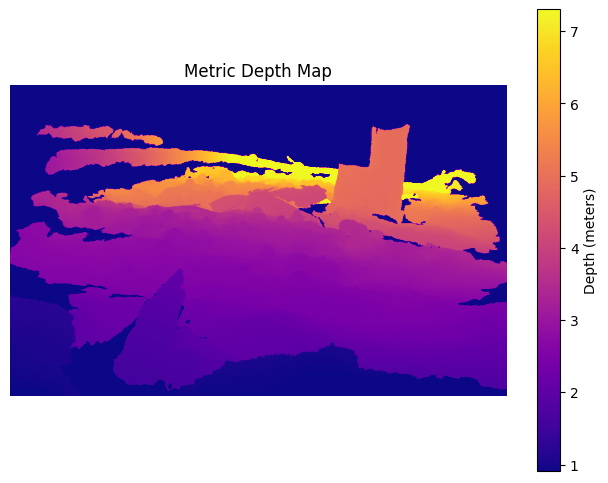} &
        \includegraphics[width=0.115\textwidth]{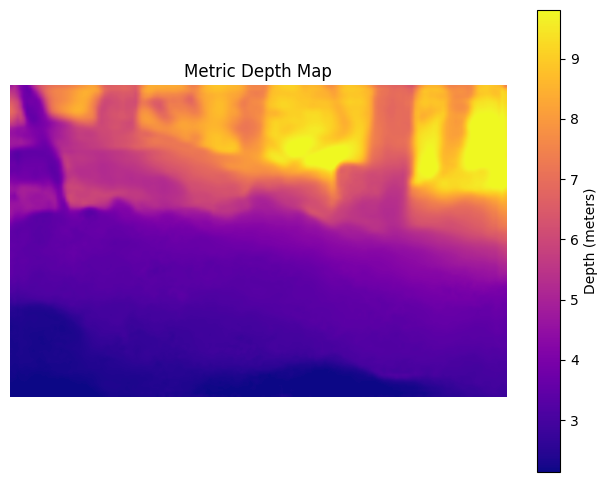} &
        \includegraphics[width=0.115\textwidth]{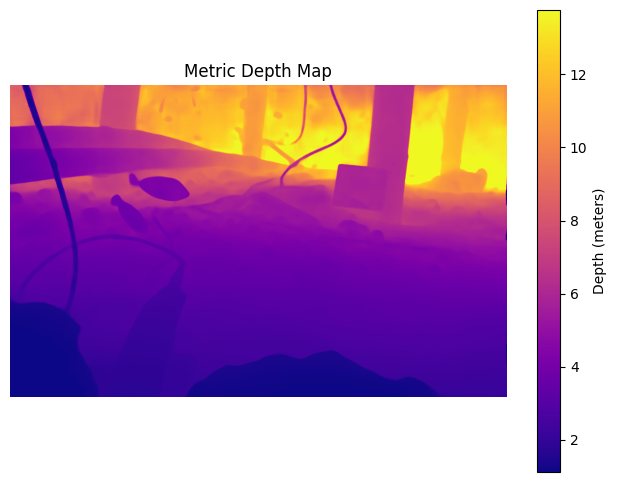} &
        \includegraphics[width=0.115\textwidth]{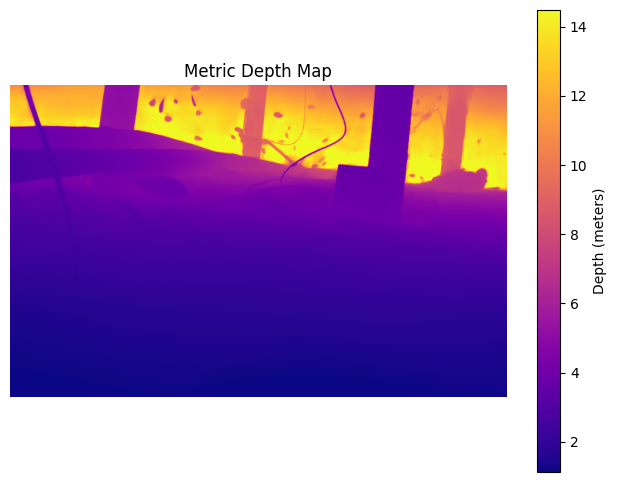} &
        \includegraphics[width=0.115\textwidth]{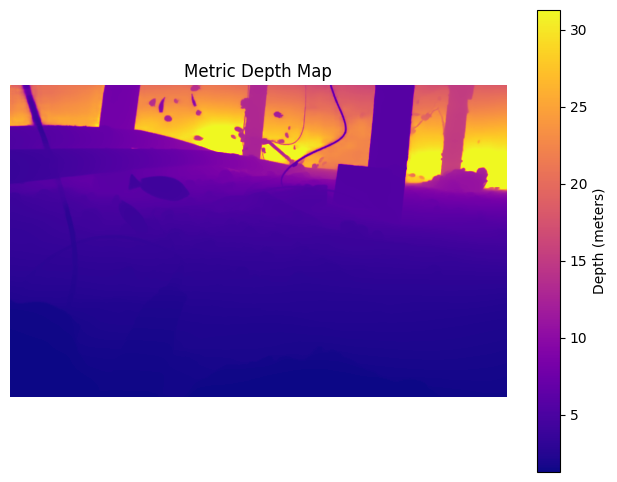} &
        \includegraphics[width=0.115\textwidth]{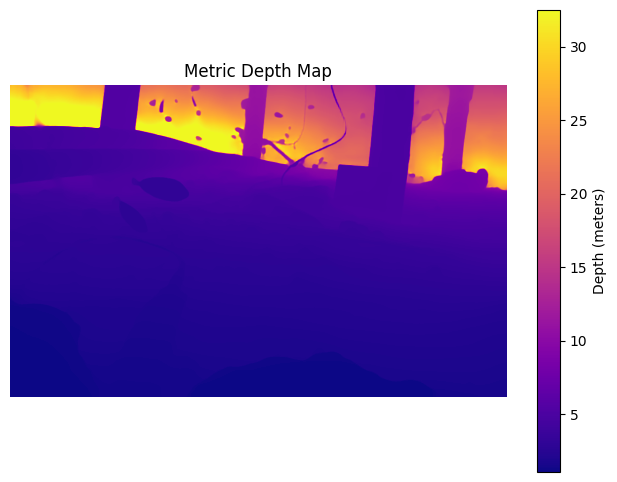} &
        \includegraphics[width=0.115\textwidth]{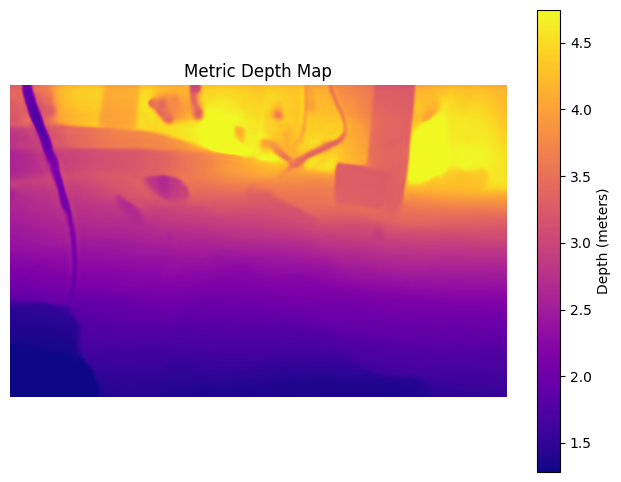} \\
    \end{tabular}

    \vspace{6pt}

    \begin{tabular}{cccccccc}
        \multicolumn{8}{c}{\textbf{SQUID: Satil}} \\
        \textbf{RGB} & \textbf{GT} & \textbf{UW-Depth$^{\dagger}$} & \textbf{Metric3D-L$^{\dagger}$} & 
        \textbf{DA V2-L$^{\dagger}$} & \textbf{Depth Pro} & \textbf{UniDepth-L} & \textbf{ZoeDepth} \\
        \includegraphics[width=0.115\textwidth]{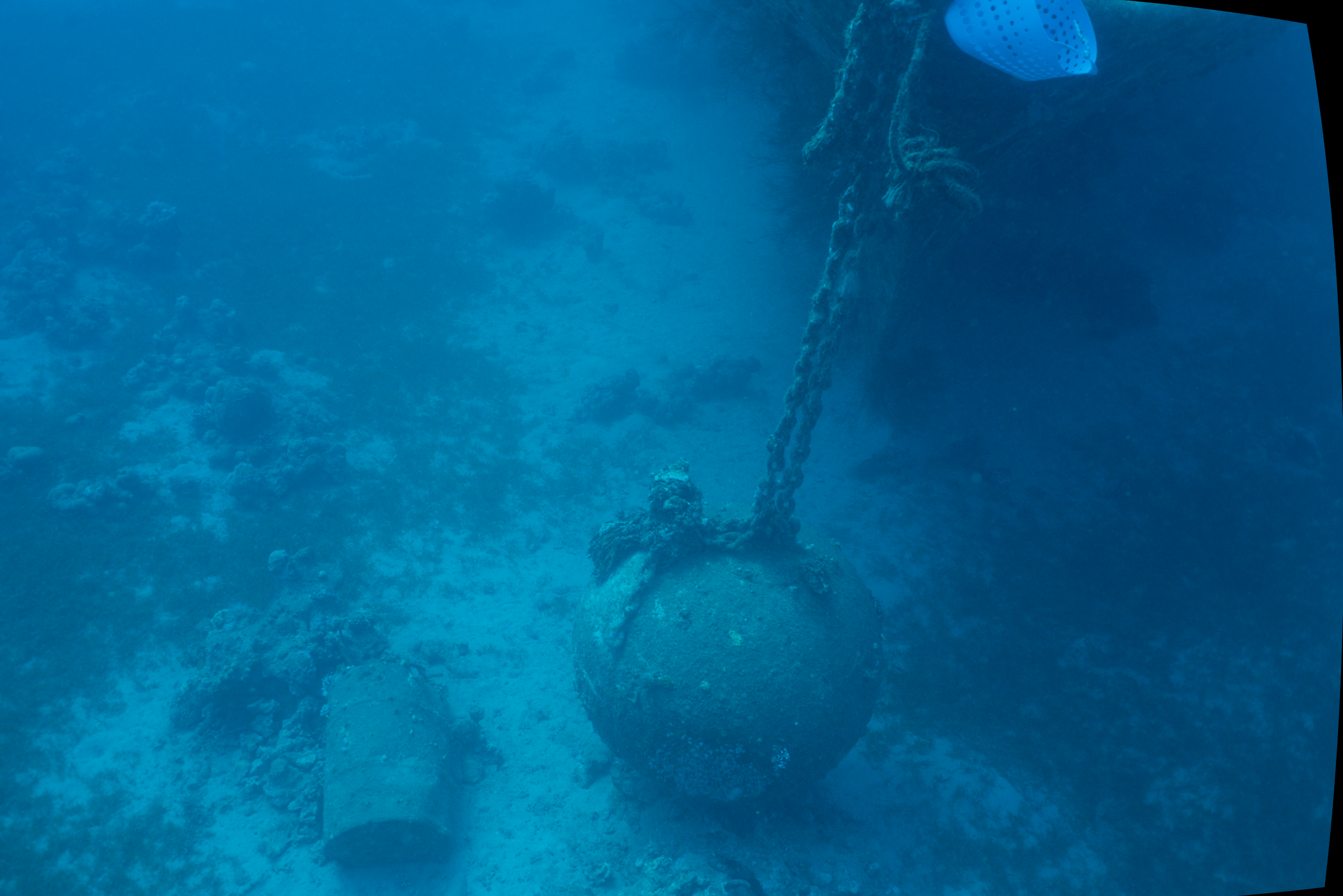} &
        \includegraphics[width=0.115\textwidth]{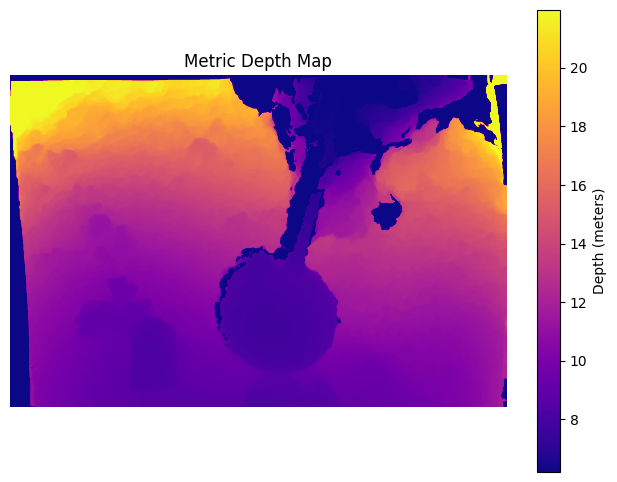} &
        \includegraphics[width=0.115\textwidth]{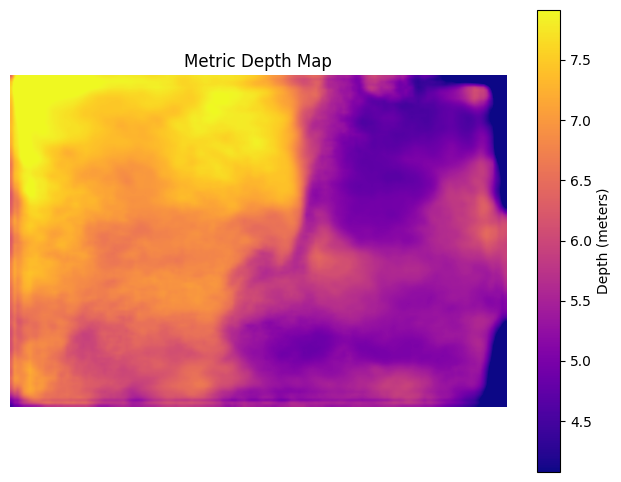} &
        \includegraphics[width=0.115\textwidth]{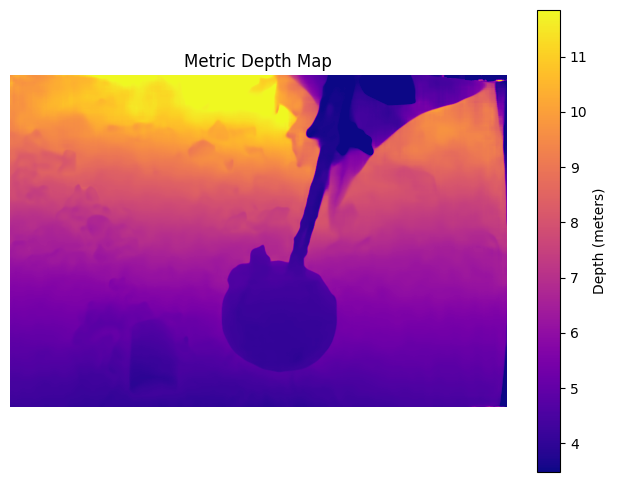} &
        \includegraphics[width=0.115\textwidth]{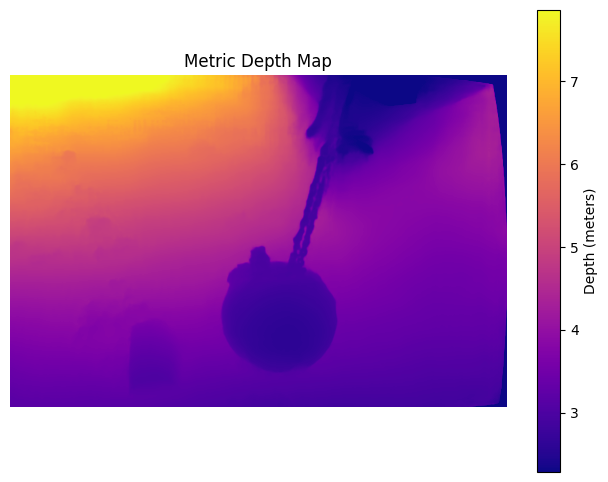} &
        \includegraphics[width=0.115\textwidth]{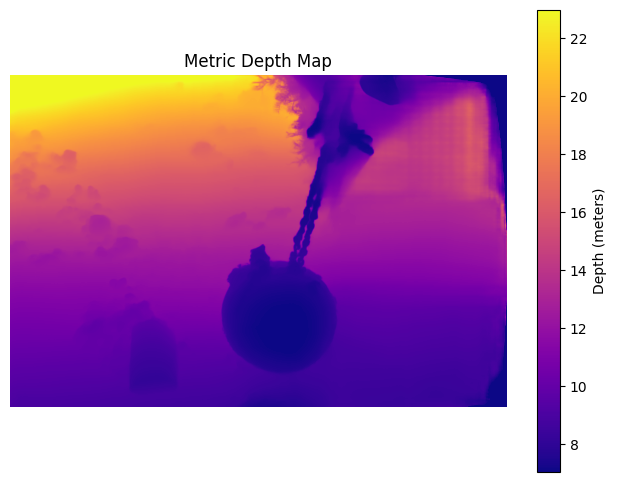} &
        \includegraphics[width=0.115\textwidth]{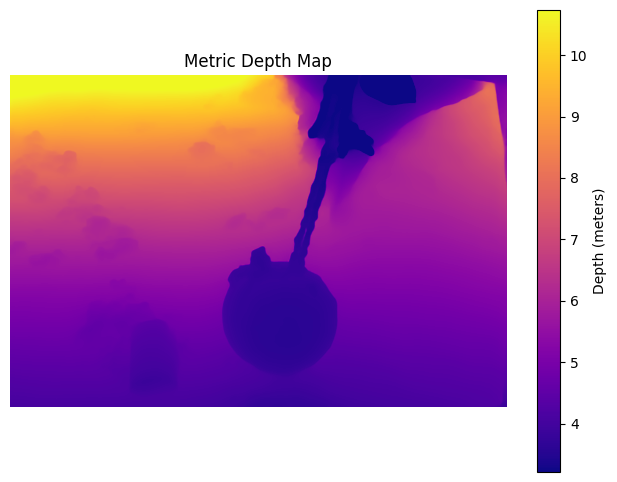} &
        \includegraphics[width=0.115\textwidth]{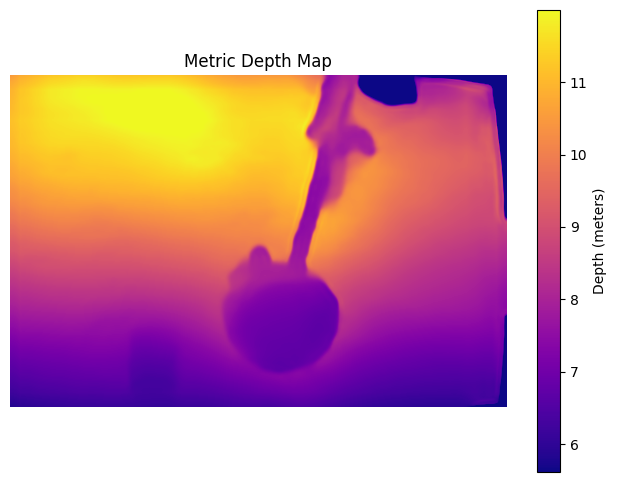} \\
    \end{tabular}

    \vspace{6pt}

    \begin{tabular}{cccccccc}
        \multicolumn{8}{c}{\textbf{SQUID: Nachsolim}} \\
        \textbf{RGB} & \textbf{GT} & \textbf{UW-Depth$^{\dagger}$} & \textbf{Metric3D-L$^{\dagger}$} & 
        \textbf{DA V2-L$^{\dagger}$} & \textbf{Depth Pro} & \textbf{UniDepth-L} & \textbf{ZoeDepth} \\
        \includegraphics[width=0.115\textwidth]{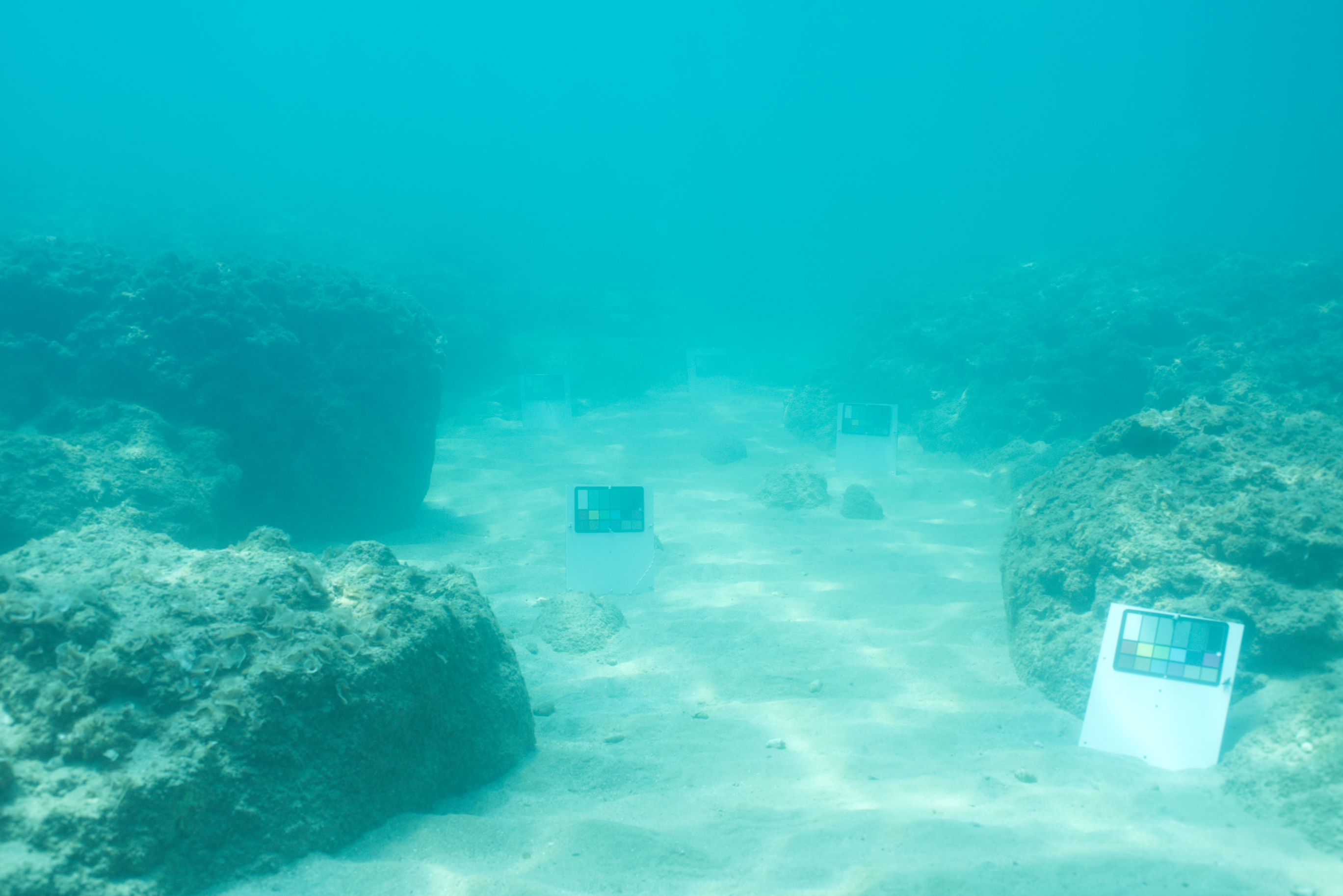} &
        \includegraphics[width=0.115\textwidth]{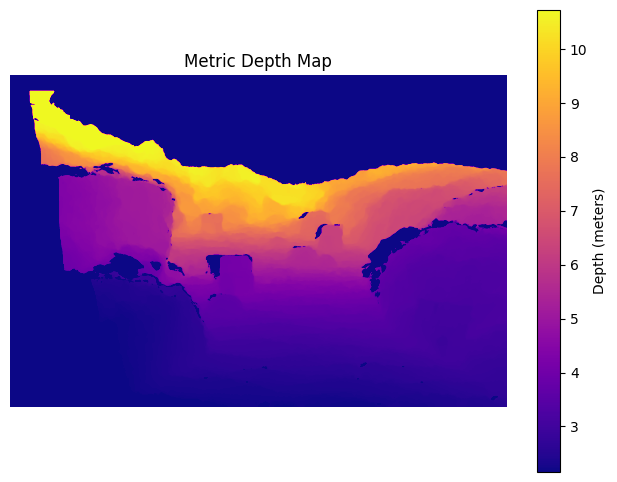} &
        \includegraphics[width=0.115\textwidth]{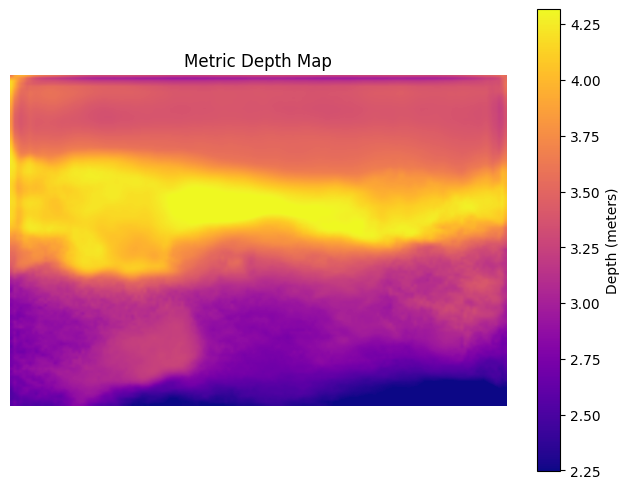} &
        \includegraphics[width=0.115\textwidth]{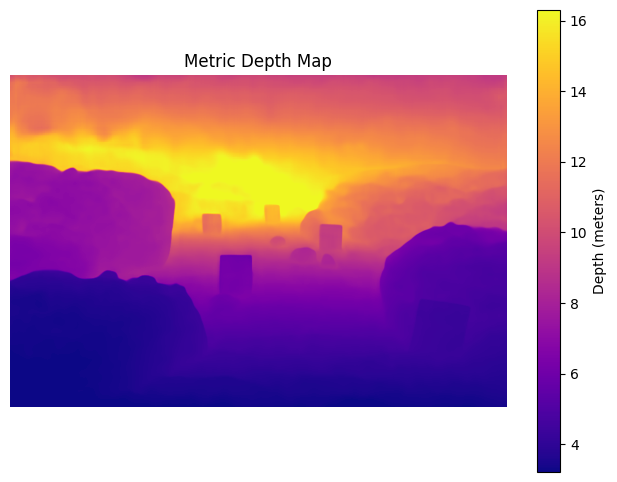} &
        \includegraphics[width=0.115\textwidth]{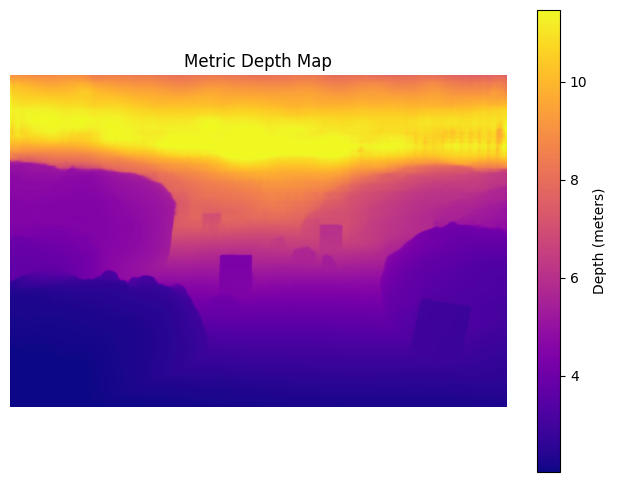} &
        \includegraphics[width=0.115\textwidth]{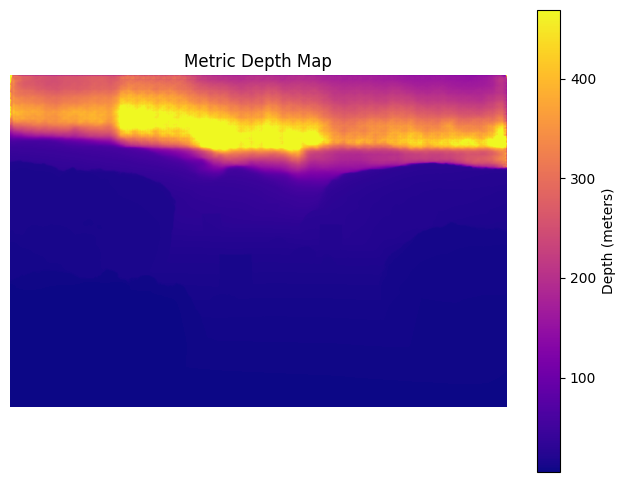} &
        \includegraphics[width=0.115\textwidth]{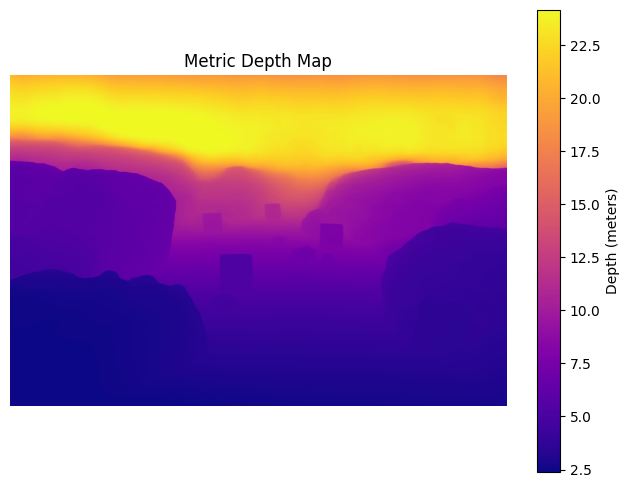} &
        \includegraphics[width=0.115\textwidth]{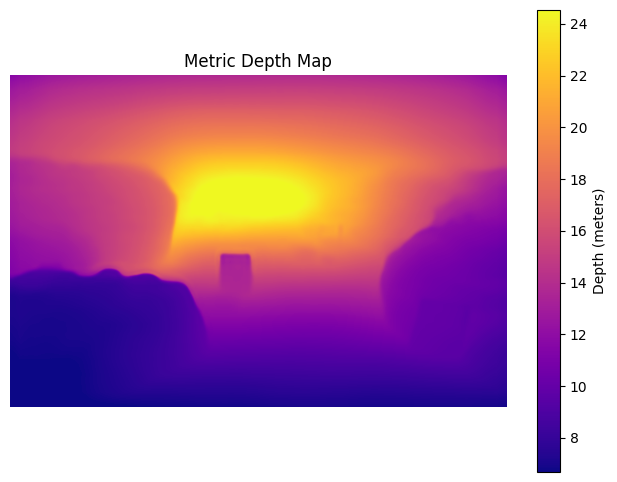} \\
    \end{tabular}

    \vspace{6pt}

    \begin{tabular}{cccccccc}
        \multicolumn{8}{c}{\textbf{SQUID: Michmoret}} \\
        \textbf{RGB} & \textbf{GT} & \textbf{UW-Depth$^{\dagger}$} & \textbf{Metric3D-L$^{\dagger}$} & 
        \textbf{DA V2-L$^{\dagger}$} & \textbf{Depth Pro} & \textbf{UniDepth-L} & \textbf{ZoeDepth} \\
        \includegraphics[width=0.115\textwidth]{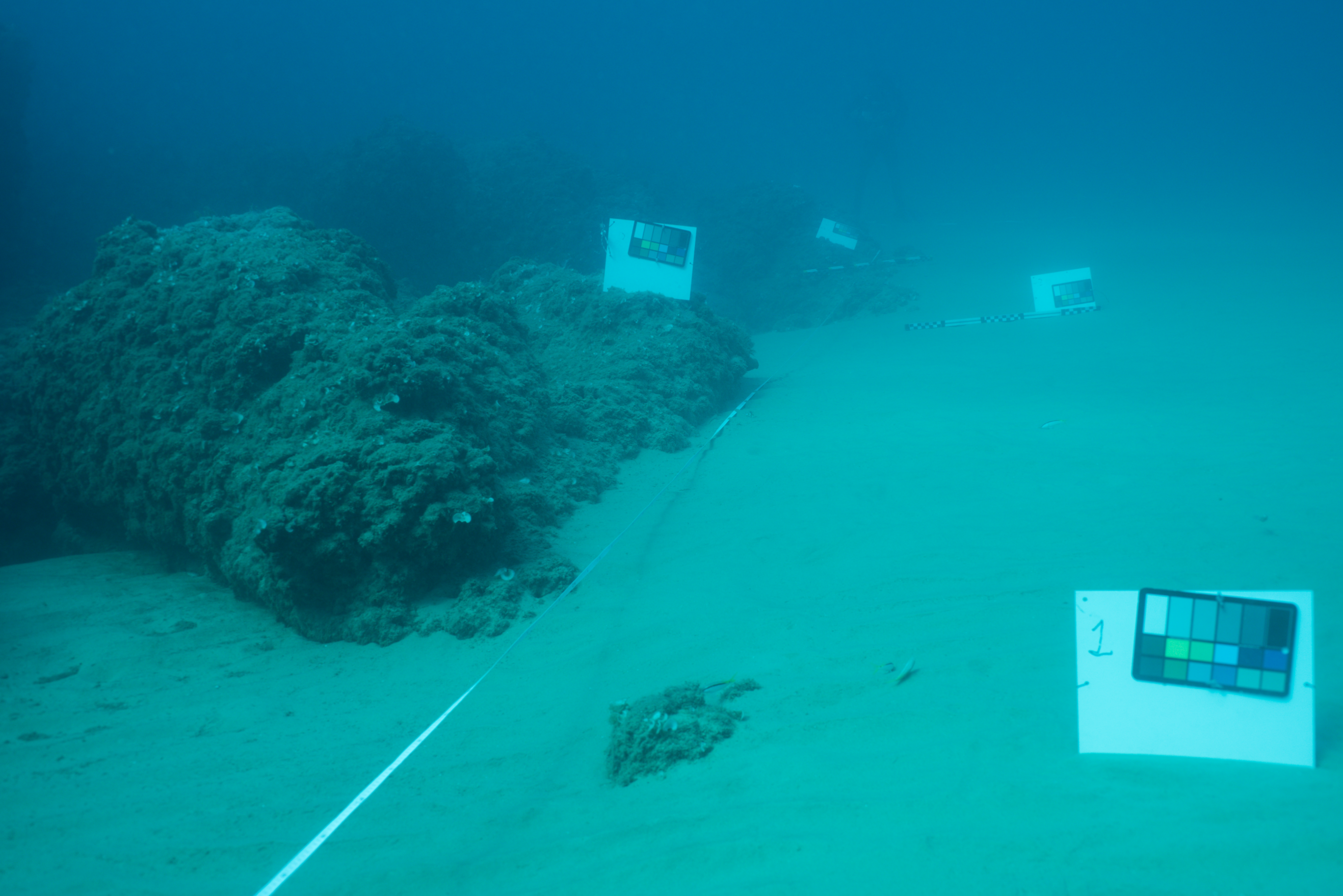} &
        \includegraphics[width=0.115\textwidth]{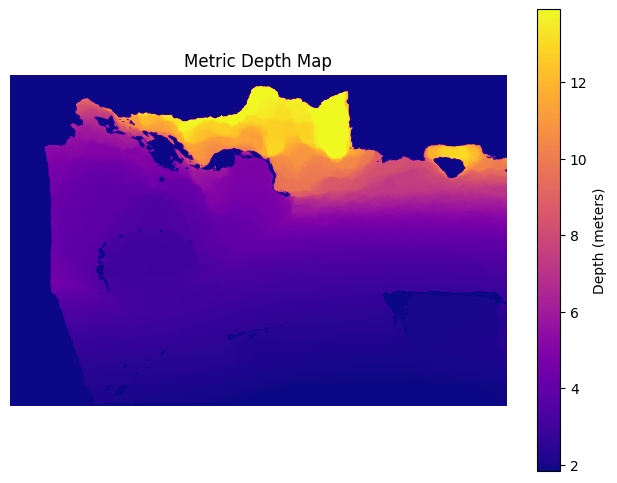} &
        \includegraphics[width=0.115\textwidth]{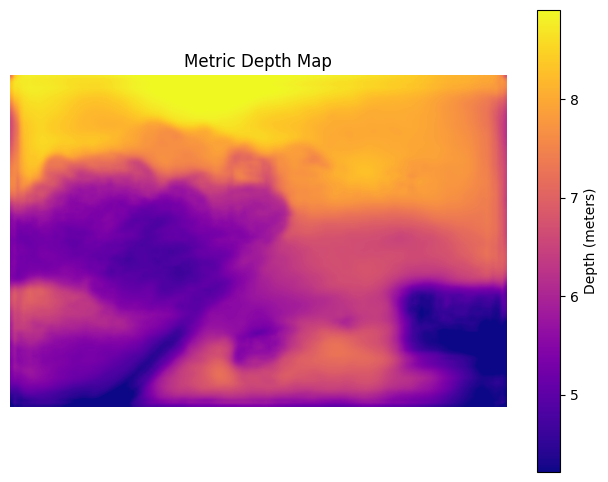} &
        \includegraphics[width=0.115\textwidth]{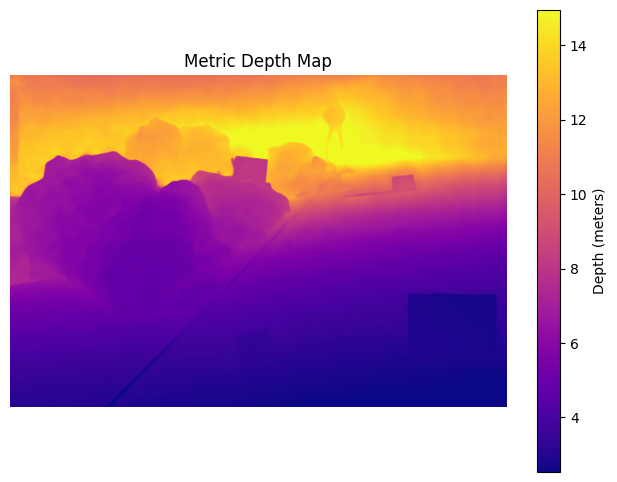} &
        \includegraphics[width=0.115\textwidth]{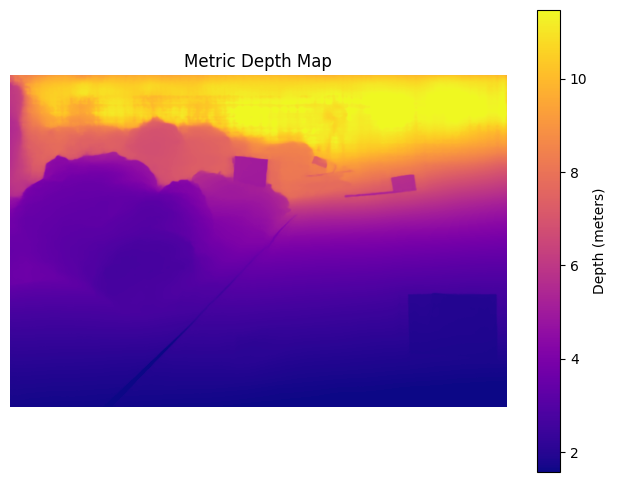} &
        \includegraphics[width=0.115\textwidth]{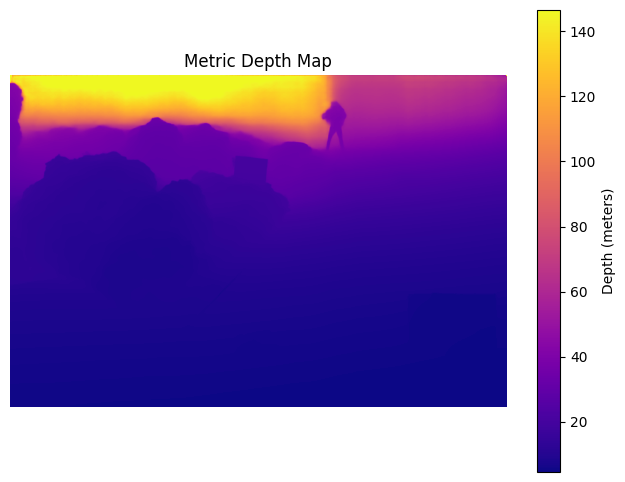} &
        \includegraphics[width=0.115\textwidth]{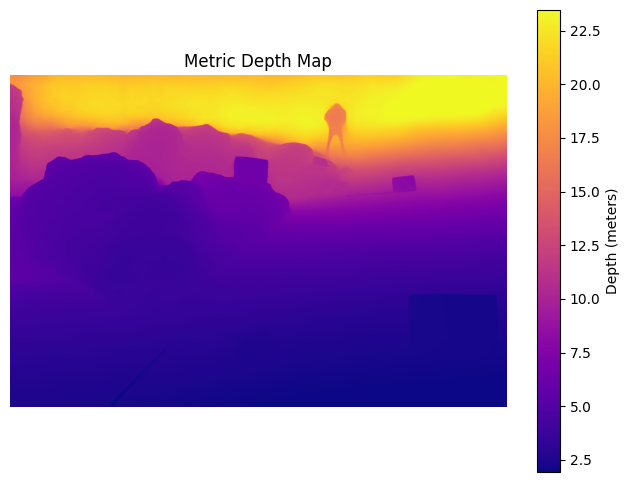} &
        \includegraphics[width=0.115\textwidth]{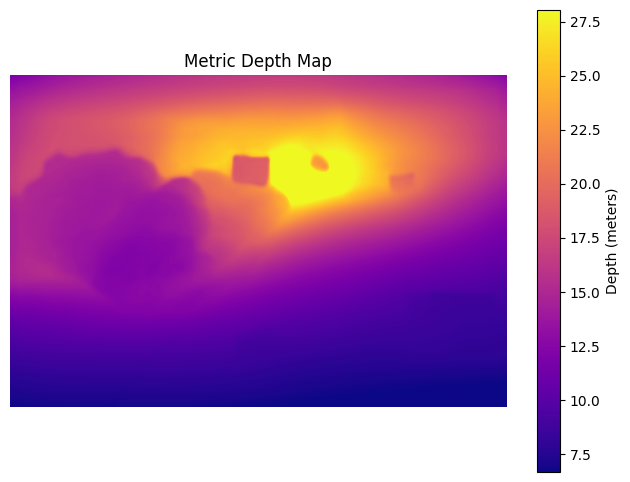} \\
    \end{tabular}

    \vspace{6pt}

    \begin{tabular}{cccccccc}
        \multicolumn{8}{c}{\textbf{SQUID: Katzaa}} \\
        \textbf{RGB} & \textbf{GT} & \textbf{UW-Depth$^{\dagger}$} & \textbf{Metric3D-L$^{\dagger}$} & 
        \textbf{DA V2-L$^{\dagger}$} & \textbf{Depth Pro} & \textbf{UniDepth-L} & \textbf{ZoeDepth} \\
        \includegraphics[width=0.115\textwidth]{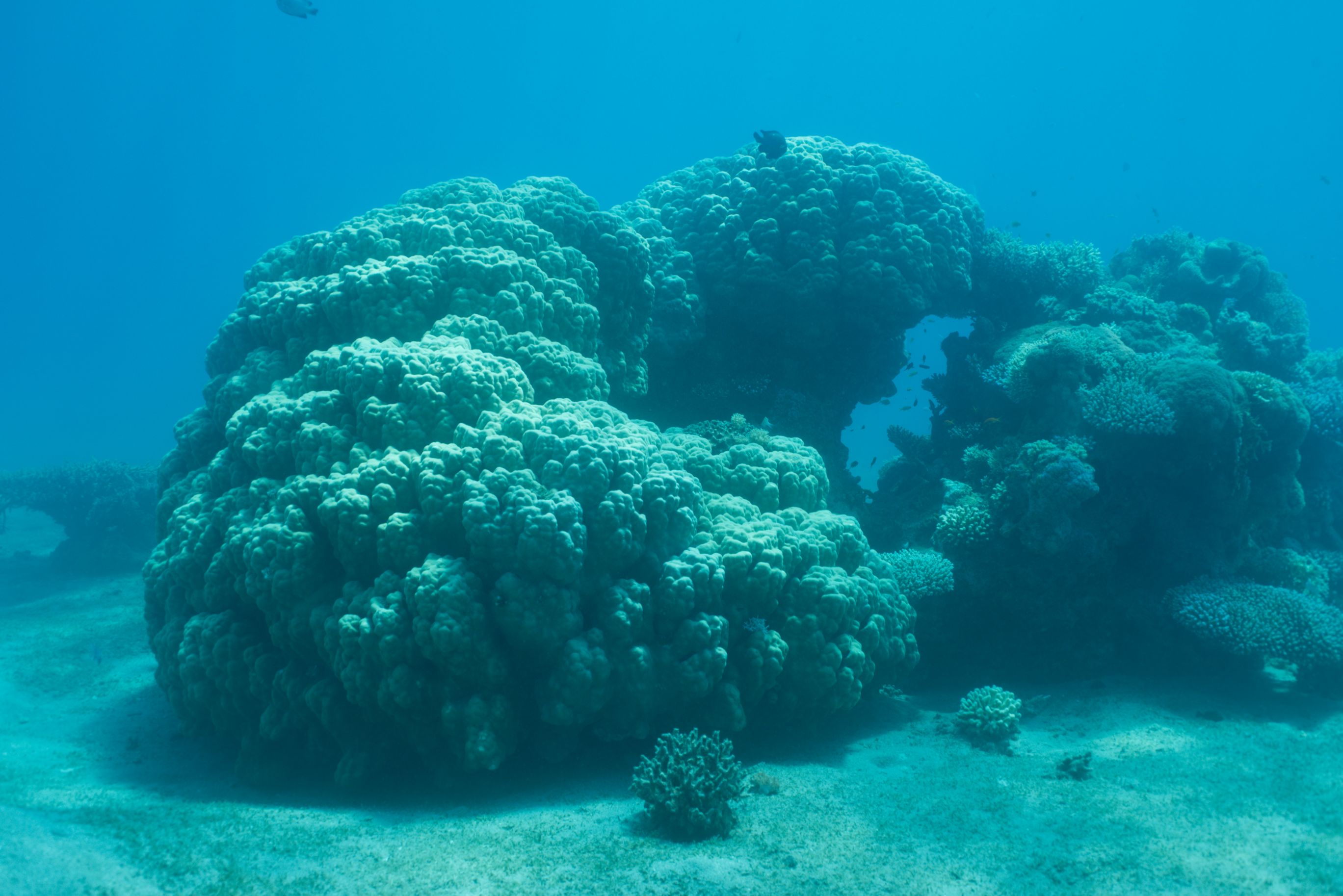} &
        \includegraphics[width=0.115\textwidth]{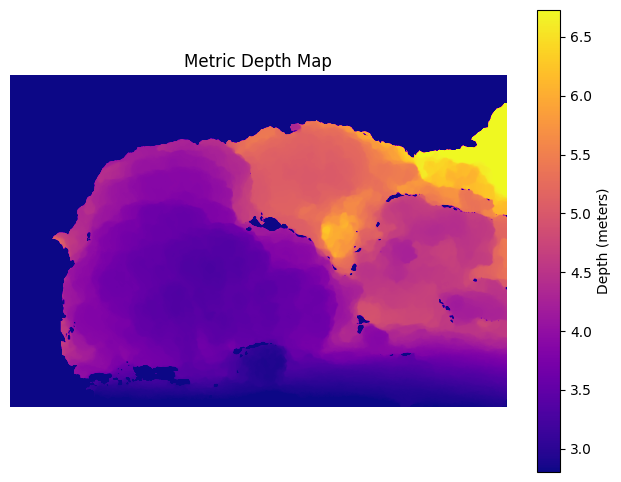} &
        \includegraphics[width=0.115\textwidth]{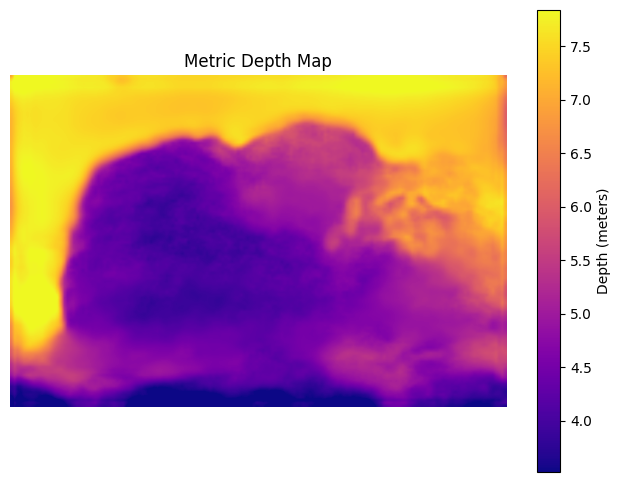} &
        \includegraphics[width=0.115\textwidth]{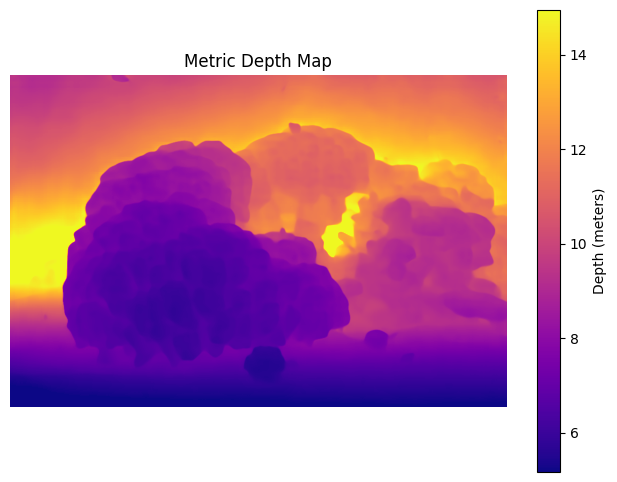} &
        \includegraphics[width=0.115\textwidth]{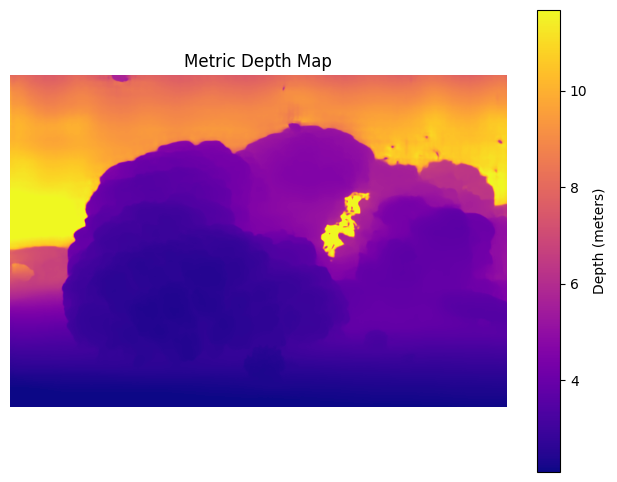} &
        \includegraphics[width=0.115\textwidth]{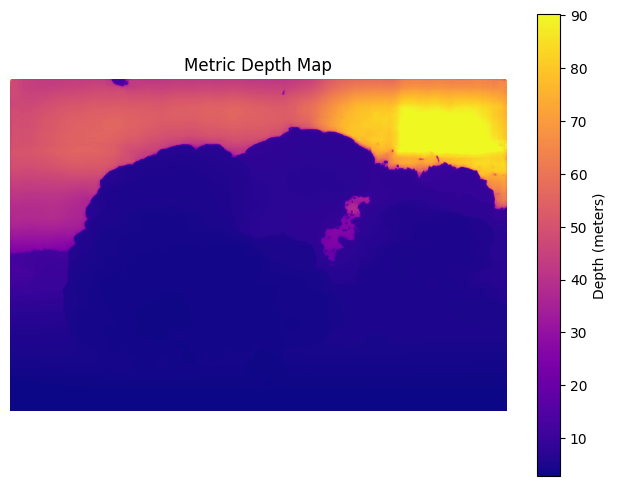} &
        \includegraphics[width=0.115\textwidth]{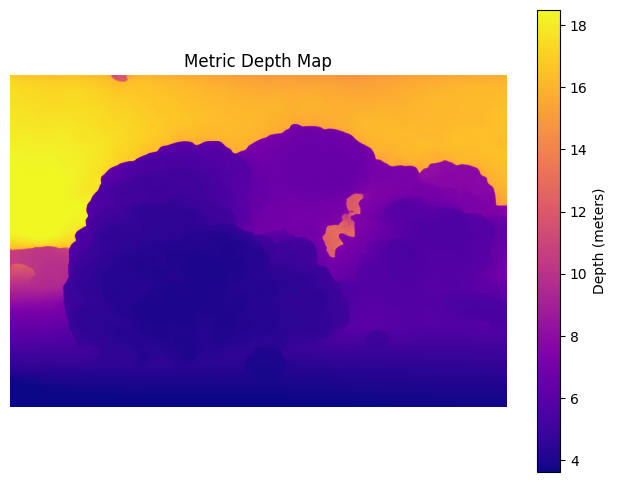} &
        \includegraphics[width=0.115\textwidth]{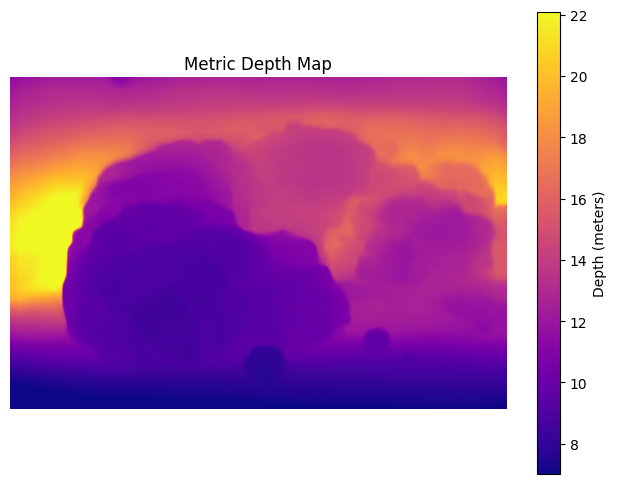} \\
    \end{tabular}

    \caption{
    \scriptsize
    Qualitative comparisons on 6 underwater scenes from two real-world underwater datasets: FLSea and SQUID. Each group shows the RGB input, ground-truth (GT), UW-Depth$^{\dagger}$ \cite{ebner2023metricallyscaledmonoculardepth}, Metric3D V2$^{\dagger}$ \cite{10638254}, Depth Anything V2 (ViT-L)$^{\dagger}$ \cite{yang2024depthv2}, Depth Pro \cite{bochkovskii2025depthprosharpmonocular}, UniDepth V2 (ViT-L) \cite{piccinelli2025unidepthv2universalmonocularmetric}, and ZoeDepth \cite{bhat2023zoedepthzeroshottransfercombining}. UniDepth V2 (ViT-L) \cite{piccinelli2025unidepthv2universalmonocularmetric} consistently produces the most accurate metric depth maps with fine edge details across all datasets. Depth Anything V2 (ViT-L) \cite{yang2024depthv2} also performs competitively, particularly in near-range scenes (\textless 20 meters), but shows softer boundaries and more artifacts at longer ranges. Metric3D V2 (ViT-L) \cite{10638254} outputs visually sharpest maps, but its metric depth scale accuracy is unclear due to the lack of its quantitative results in our study. Depth Pro \cite{bochkovskii2025depthprosharpmonocular} is affected the most by texture-less regions in the background, leading to poor structural recovery for the foreground. ZoeDepth \cite{bhat2023zoedepthzeroshottransfercombining} and UW-Depth \cite{ebner2023metricallyscaledmonoculardepth} both underperform due to their limited model capacity and training domain scope.
    }
    \label{fig:qualitative_results_all}
\end{figure*}

\begin{figure*}[ht]
    \centering
    \renewcommand{\arraystretch}{1.0}
    \setlength{\tabcolsep}{1pt}
    \scriptsize

    \begin{tabular}{cccccccc}
        \multicolumn{4}{c}{\textbf{FLSea: Canyon}} &
        \multicolumn{4}{c}{\textbf{FLSea: Red sea}} \\
        \textbf{RGB} & \textbf{GT} & \textbf{Baseline} & \textbf{Fine-tuned (Ours)} &
        \textbf{RGB} & \textbf{GT} & \textbf{Baseline} & \textbf{Fine-tuned (Ours)} \\        \includegraphics[width=0.11\textwidth]{sec/figures/flsea-1/image_gt.png} &
        \includegraphics[width=0.11\textwidth]{sec/figures/flsea-1/depth_gt.png} &
        \includegraphics[width=0.11\textwidth]{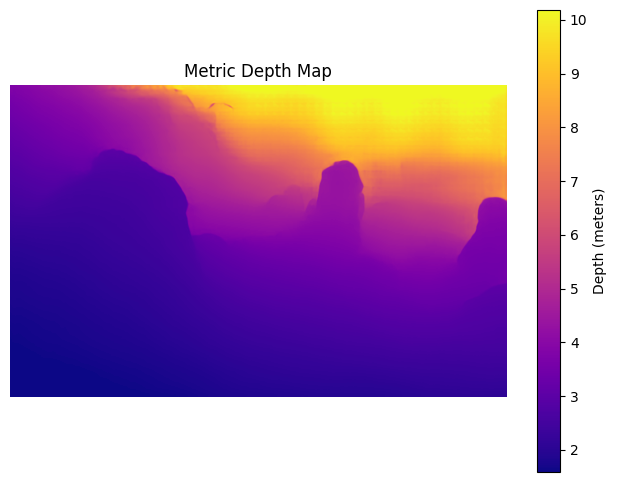} &
        \includegraphics[width=0.11\textwidth]{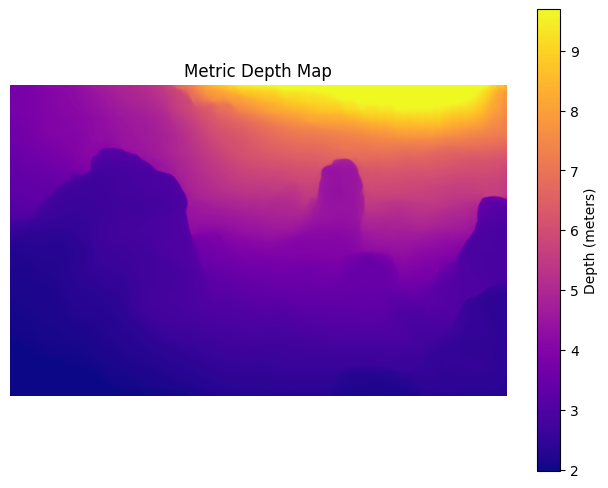} &
        \includegraphics[width=0.11\textwidth]{sec/figures/flsea-2/image_gt.png} &
        \includegraphics[width=0.11\textwidth]{sec/figures/flsea-2/depth_gt.png} &
        \includegraphics[width=0.11\textwidth]{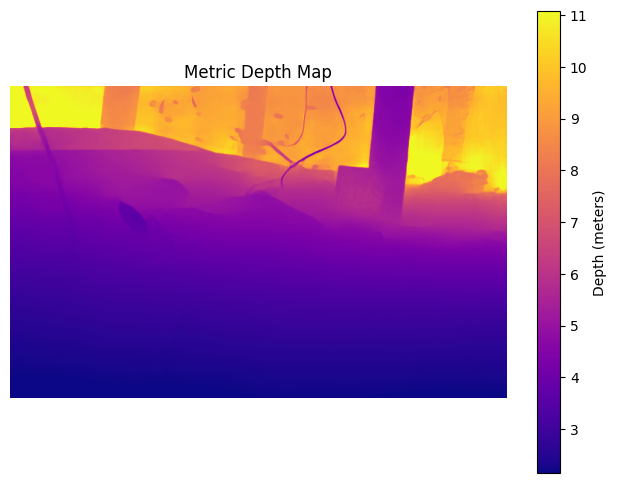} &
        \includegraphics[width=0.11\textwidth]{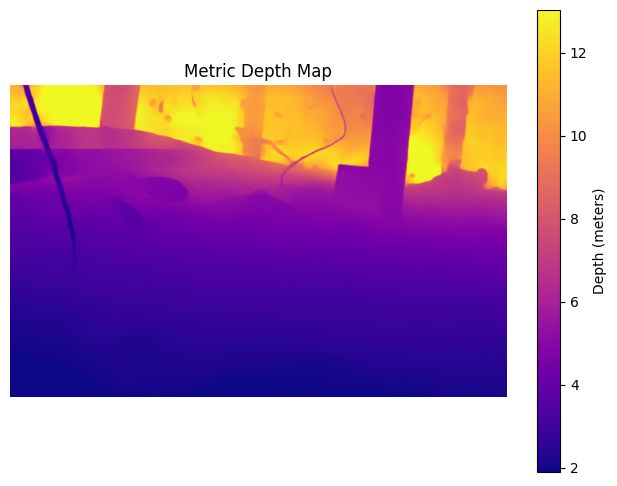}  \\
    \end{tabular}

    \vspace{6pt}

    \begin{tabular}{cccccccc}
        \multicolumn{4}{c}{\textbf{SQUID: Satil}} &
        \multicolumn{4}{c}{\textbf{SQUID: Nachsolim}} \\
        \textbf{RGB} & \textbf{GT} & \textbf{Baseline} & \textbf{Fine-tuned (Ours)} &
        \textbf{RGB} & \textbf{GT} & \textbf{Baseline} & \textbf{Fine-tuned (Ours)} \\
        \includegraphics[width=0.11\textwidth]{sec/figures/squidscene1/image_gt.png} &
        \includegraphics[width=0.11\textwidth]{sec/figures/squidscene1/depth_gt.png} &
        \includegraphics[width=0.11\textwidth]{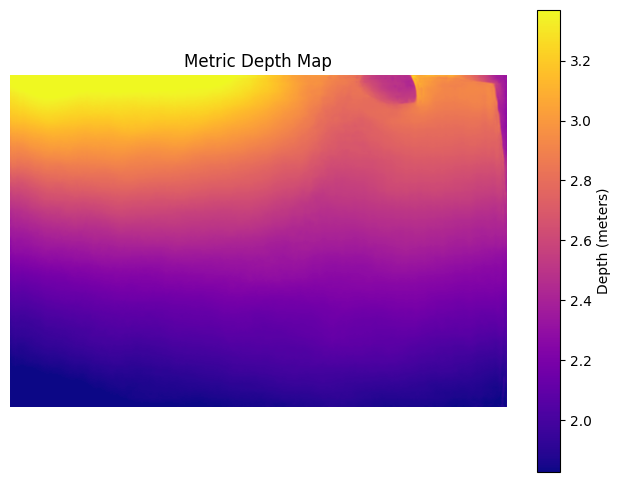} &
        \includegraphics[width=0.11\textwidth]{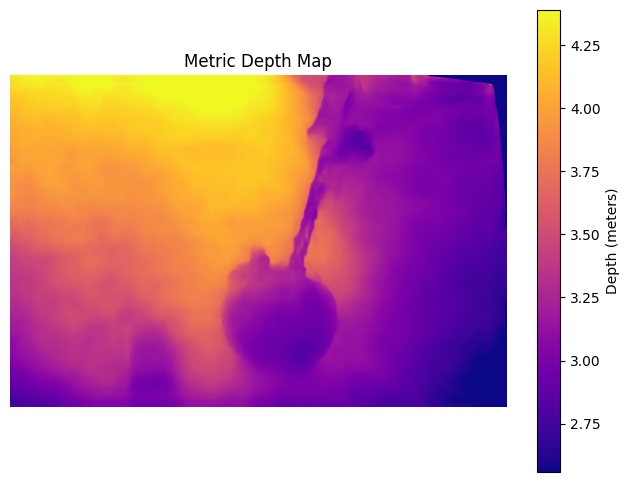} &
        \includegraphics[width=0.11\textwidth]{sec/figures/squidscene2/image_gt.png} &
        \includegraphics[width=0.11\textwidth]{sec/figures/squidscene2/depth_gt.png} &
        \includegraphics[width=0.11\textwidth]{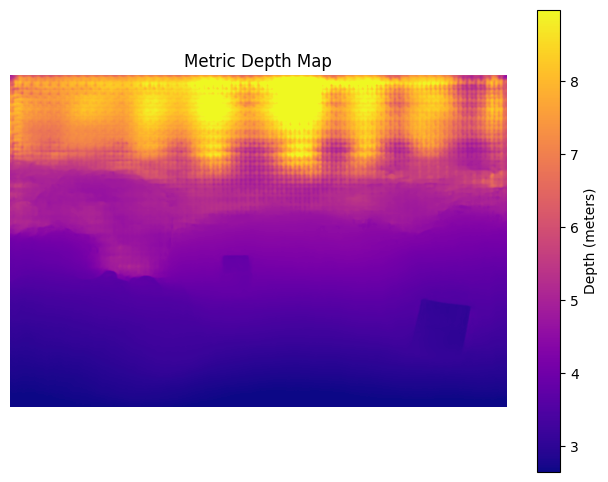} &
        \includegraphics[width=0.11\textwidth]{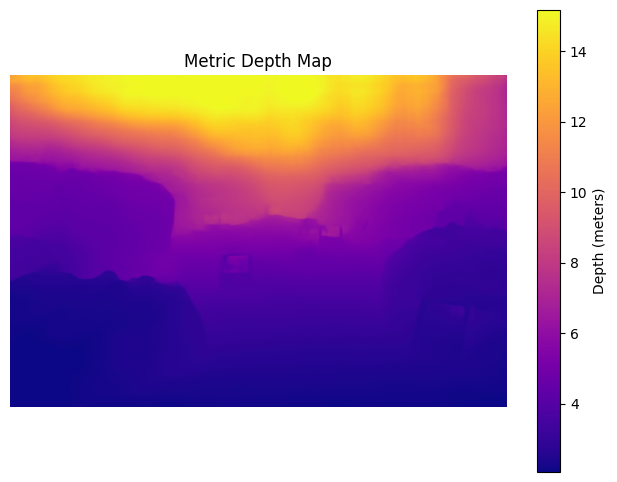} \\
    \end{tabular}

    \vspace{6pt}

    \begin{tabular}{cccccccc}
        \multicolumn{4}{c}{\textbf{SQUID: Michmoret}} &
        \multicolumn{4}{c}{\textbf{SQUID: Katzaa}} \\
        \textbf{RGB} & \textbf{GT} & \textbf{Baseline} & \textbf{Fine-tuned (Ours)} &
        \textbf{RGB} & \textbf{GT} & \textbf{Baseline} & \textbf{Fine-tuned (Ours)} \\
        \includegraphics[width=0.11\textwidth]{sec/figures/squidscene3/image_gt.png} &
        \includegraphics[width=0.11\textwidth]{sec/figures/squidscene3/depth_gt.png} &
        \includegraphics[width=0.11\textwidth]{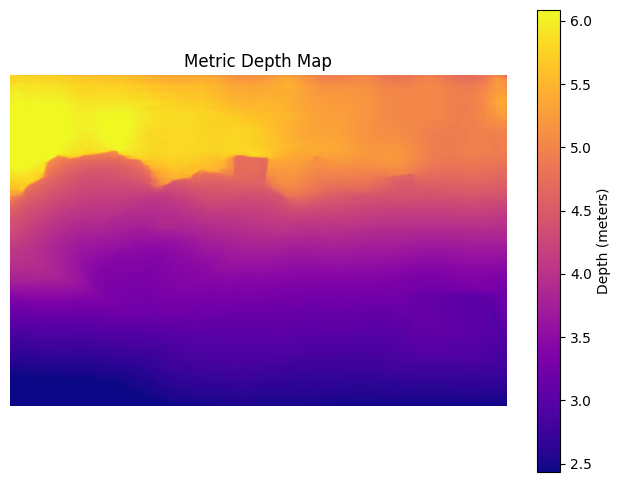} &
        \includegraphics[width=0.11\textwidth]{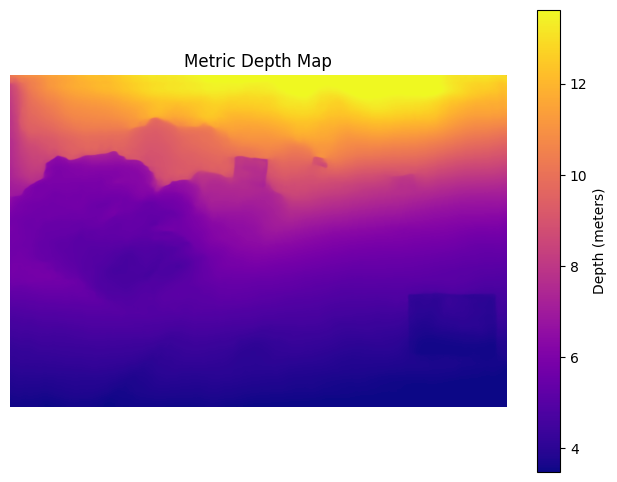} &
        \includegraphics[width=0.11\textwidth]{sec/figures/squidscene4/image_gt.png} &
        \includegraphics[width=0.11\textwidth]{sec/figures/squidscene4/depth_gt.png} &
        \includegraphics[width=0.11\textwidth]{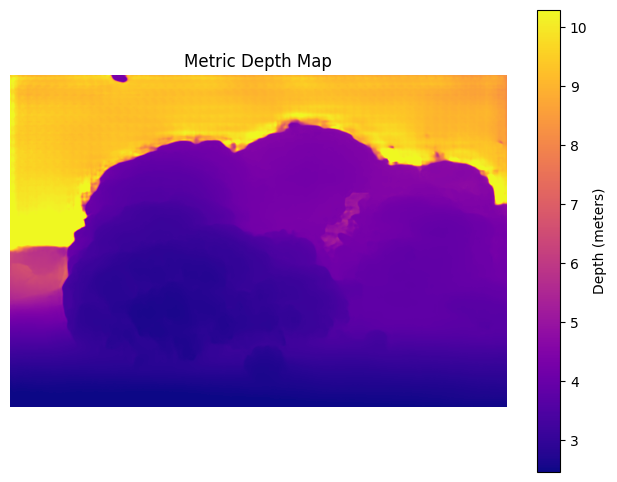} &
        \includegraphics[width=0.11\textwidth]{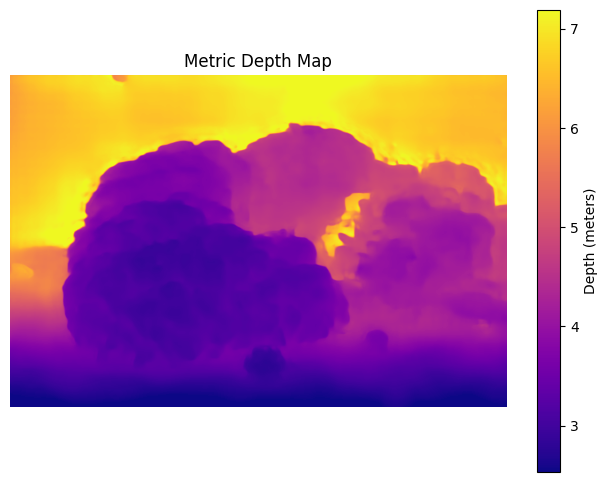}  \\
    \end{tabular}

    \caption{Qualitative comparison between the baseline (Depth Anything V2 ViT-S \cite{yang2024depthv2}) and our fine-tuned model using synthetic underwater data on six scenes from two real-world datasets: FLSea \cite{randall2023flsea} and SQUID \cite{berman2018underwater}. Each group shows the RGB input, ground-truth (GT), zero-shot baseline prediction, and prediction after fine-tuning. Our synthetically fine-tuned models produce much sharper depth boundaries and a more accurate metric depth scale with improved robustness for adapting underwater domain than the baseline models across all scenes, especially in turbid and low-contrast regions seen in the SQUID\cite{berman2018underwater} dataset with high scattering and color distortion. 
    }
    \label{fig:qualitative_finetune}
\end{figure*}

We present qualitative comparisons of predicted depth maps on representative scenes from FLSea \cite{randall2023flsea} and SQUID \cite{berman2018underwater}. \Cref{fig:qualitative_results_all} showcases zero-shot performance across all benchmarked models using their strongest performing variants, highlighting differences in detail boundary preservation and robustness to underwater artifacts. \Cref{fig:qualitative_finetune} compares the Depth Anything V2 (ViT-S) \cite{yang2024depthv2} baseline with our fine-tuned model, demonstrating the visual impact and effectiveness of synthetic-to-real domain adaptation \cite{10161531, 10048777}.

%% file: sec/5_discussion.tex
\section{Discussion}
\label{sec:discussion}

\subsection{Zero-Shot Model Performance}
Our results demonstrate that general-purpose monocular depth models—such as Depth Anything V2 \cite{yang2024depthv2} and UniDepth V2 \cite{piccinelli2025unidepthv2universalmonocularmetric}—perform reasonably well on underwater imagery in a zero-shot setting. However, performance varies significantly across underwater conditions. On datasets with clearer water and narrower depth ranges (e.g., FLSea \cite{randall2023flsea}), models retain moderate accuracy. In contrast, performance degrades in more turbid or visually degraded scenes, such as SQUID \cite{berman2018underwater}, with higher \textit{AbsRel} errors and lower $\delta_1$ accuracy \cite{eigen2014depthmappredictionsingle, he2025distilldepthdistillationcreates}. Furthermore, models like ZoeDepth \cite{bhat2023zoedepthzeroshottransfercombining} and Depth Pro \cite{bochkovskii2025depthprosharpmonocular}, while robust on benchmarks like NYU-V2 \cite{couprie2013indoor} and KITTI \cite{geiger2013vision}, experience a significant performance drop when applied to underwater settings. This highlights the difficulty of transferring terrestrial-trained models to underwater domains without explicit adaptation. This is further supported by the performance of the UW-Depth \cite{ebner2023metricallyscaledmonoculardepth} model trained on real-world underwater data with a compact MobileNet V2 backbone \cite{sandler2019mobilenetv2invertedresidualslinear}, yet achieving greater performance than the other larger models. Overall, UniDepth V2 \cite{piccinelli2025unidepthv2universalmonocularmetric} performs exceptionally well on the FLSea \cite{randall2023flsea} dataset, achieving $\delta_1$ accuracy above 90\% and the lowest \textit{AbsRel} errors across all benchmarked models \cite{eigen2014depthmappredictionsingle, he2025distilldepthdistillationcreates}.

\subsection{Effectiveness of Synthetic Fine-Tuning}
Fine-tuning Depth Anything V2 (ViT-S) \cite{yang2024depthv2} on our physics-based synthetic underwater dataset yields consistent improvements across real-world test sets. Quantitatively, fine-tuning reduces absolute errors and improves threshold accuracy on all SQUID \cite{berman2018underwater} scenes and most FLSea \cite{randall2023flsea} subsets. Qualitatively (\cref{fig:qualitative_finetune}), the model learns to recover better structural details, suppresses noisy predictions in low-contrast areas, and improves depth continuity in visually degraded regions. This underscores the utility of synthetic domain adaptation for underwater vision tasks where collecting metric-labeled real data is costly or infeasible \cite{9570386, 10048777, raveendran2021underwater}.

\subsection{Qualitative Observations}
Across zero-shot and fine-tuned visual depth map comparisons (\cref{fig:qualitative_results_all} and \cref{fig:qualitative_finetune}), we observe:
\begin{itemize}

    \item \textbf{UniDepth V2} \cite{piccinelli2025unidepthv2universalmonocularmetric}: Among all evaluated models, UniDepth V2 \cite{piccinelli2025unidepthv2universalmonocularmetric} consistently delivers the best overall performance across all benchmarks. It produces accurate metric depth maps with fine structural details and strong generalization ability to varying underwater conditions.  

    \item \textbf{Depth Anything V2 (fine-tuned)} \cite{yang2024depthv2}: The baseline Depth Anything V2 \cite{yang2024depthv2} ranks just behind UniDepth V2 \cite{piccinelli2025unidepthv2universalmonocularmetric}, showing strong metric scale accuracy but slightly less precise boundary detail. After fine-tuning the ViT-S variant on our synthetic underwater dataset, the model exhibits improved structural consistency, sharper depth boundaries, and more accurate metric depth predictions.
    
    \item \textbf{UW-Depth} \cite{ebner2023metricallyscaledmonoculardepth}: While the compact backbone design limits spatial resolution and leads to loss of structural detail in predicted depth maps, the model still provides reasonably accurate metric scale predictions, especially in scenes similar to its training distribution.
    
    \item \textbf{Metric3D V2} \cite{10638254}: The ViT-L variant of this model visually performs on par with UniDepth. However, it is difficult to confirm its exact performance without any quantitative evaluation, highlighting one of our key limitations. 
    
    \item \textbf{ZoeDepth \cite{bhat2023zoedepthzeroshottransfercombining} and Depth Pro \cite{bochkovskii2025depthprosharpmonocular}}: While producing relatively smooth depth maps with decent edge boundary preservation, these models tend to have more inaccuracies in metric scale prediction compared to others.
    
\end{itemize}

\subsection{Model Trade-offs}
Transformer-based models with larger encoders (e.g., ViT-L) typically offer better zero-shot accuracy and generalization. However, they also introduce extra computational overhead, which can make them less practical for real-time deployment on embedded systems \cite{sandler2019mobilenetv2invertedresidualslinear}. Lightweight models such as UW-Depth \cite{ebner2023metricallyscaledmonoculardepth} provide a viable speed-accuracy trade-off, which suggests a gap between compact model efficiency and underwater robustness.

\subsection{Limitations and Future Work}
While synthetic fine-tuning notably improves performance, several challenges remain:
\begin{itemize}
    \item Performance significantly degrades in extreme conditions, such as highly turbid water, low-light scenes, or regions with texture-less backgrounds.
    \item We only fine-tuned Depth Anything V2 (ViT-S) \cite{yang2024depthv2}; other backbone variants and models (e.g., UniDepth \cite{piccinelli2025unidepthv2universalmonocularmetric}) could also benefit from synthetic adaptation.
    \item Our synthetic dataset is limited to Hypersim-based indoor geometry \cite{roberts:2021}; future work could incorporate more diverse 3D structures and scene ranges. Also, simulating additional real-world underwater phenomena with stable diffusion and using unlabeled underwater images for self-supervised training could enhance model generalization for the underwater setting as well \cite{Zhang_2024_CVPR}.
\end{itemize}
Overall, our benchmark highlights both the promise and limitations of monocular depth estimation in underwater conditions and provides understanding for further exploration in data simulation, cross-model adaptation, and real-world deployment in future research.

%% file: sec/6_conclusion.tex
\section{Conclusion}
\label{sec:conclusion}

In this work, we presented a comprehensive benchmark of monocular metric depth estimation models for underwater environments, comparing a diverse set of state-of-the-art general-purpose and domain-specific approaches across two challenging real-world datasets: FLSea \cite{randall2023flsea} and SQUID \cite{berman2018underwater}, each representing distinct underwater conditions in terms of visibility, depth ranges, and scene complexity. Our goal was to evaluate the models' zero-shot generalization capability and explore whether physics-based synthetic data can effectively enable underwater domain adaptation.

We developed a synthetic data generation pipeline that simulates realistic underwater RGB images from Hypersim \cite{roberts:2021} using a physics-based underwater image formation model \cite{8578801} to address the lack of large-scale, annotated high-quality real-world underwater datasets. This pipeline incorporates varying wavelength-dependent attenuation and backscattering, generating paired RGB-depth data across multiple Jerlov water types \cite{Solonenko:15}.

Our results reveal that while general-purpose models (e.g., UniDepth V2 \cite{piccinelli2025unidepthv2universalmonocularmetric}, Depth Anything V2 \cite{yang2024depthv2}, Metric3D V2 \cite{10638254}, ZoeDepth \cite{bhat2023zoedepthzeroshottransfercombining}) show moderate zero-shot performance in clear water scenes with narrow depth ranges and textural information, their accuracy drops significantly in visually degraded or turbid conditions. Among all models evaluated, UniDepth V2 \cite{piccinelli2025unidepthv2universalmonocularmetric} achieves the best zero-shot performance across all datasets, particularly in preserving metric scale and structural consistency.

We further demonstrate that fine-tuning Depth Anything V2 (ViT-S) \cite{yang2024depthv2} on our synthetic underwater dataset improves both quantitative and qualitative performance, especially in low-visibility scenarios, where the baseline fails to extract structural details. The fine-tuned model produces sharper boundaries, better depth consistency, and more accurate metric predictions, confirming the effectiveness of using synthetic data for underwater domain adaptation.

Overall, our benchmark reveals the difficulty of transferring general-purpose depth models trained with mainly terrestrial data to underwater settings and the potential of synthetic data to bridge this domain gap. Future directions include: (1) extending fine-tuning to additional backbone variants (e.g., ViT-L, MobileNetV2, etc) and models like UniDepth \cite{piccinelli2025unidepthv2universalmonocularmetric}, (2) incorporating more diverse and dynamic underwater scenes, and (3) more detailed evaluation of inference speed and accuracy trade-off to help select the appropriate model on resource-constrained platforms for real-time deployments. All of these efforts will be critical to enhance the robustness of monocular metric depth models in practical real-world underwater applications. 

%% file: sec/7_acknowledgments.tex
\section*{Acknowledgments}
I would like to thank Dr. Christopher Metzler for his guidance and feedback throughout this project, and Tianfu Wang and members of the Intelligent Sensing Laboratory at the University of Maryland, College Park, for helpful discussions and suggestions during development.